\documentclass[11pt]{article}
\usepackage[final]{acl}

\usepackage[T1]{fontenc}
\usepackage[utf8]{inputenc} 

\usepackage{times}
\usepackage{latexsym}

\usepackage{amsmath,amssymb,mathtools}
\usepackage{graphicx}
\usepackage{subcaption}

\usepackage{tikz}
\usepackage{xcolor}

\usepackage{longtable}
\usepackage[table]{xcolor}
\usepackage{cuted}   
\usepackage{caption} 

\usepackage[ruled,vlined,linesnumbered]{algorithm2e}

\usepackage[loadonly,shortlabels]{enumitem}

\usepackage{booktabs}
\usepackage{pifont}
\newcommand{\cmark}{\ding{51}} 
\newcommand{\xmark}{\ding{55}} 

\usepackage{microtype}
\usepackage{inconsolata}

\usepackage{booktabs}
\usepackage{tabularx}
\usepackage{multirow}
\usepackage{pifont}

\usepackage[most]{tcolorbox}
\usepackage{fancyvrb}
\usepackage{fvextra}
\usepackage{placeins}

\DefineVerbatimEnvironment{Prompt}{Verbatim}{
  breaklines=true,
  breakanywhere=true,
  fontsize=\footnotesize,
  baselinestretch=1,
  frame=single,
  framerule=0.3pt,
  framesep=2mm,
  xleftmargin=1mm,
  xrightmargin=1mm
}

\title{EmoMM: Benchmarking and Steering MLLM for
Multimodal Emotion Recognition under Conflict and Missingness}

\author{%
\textbf{Yueru Sun}\textsuperscript{*} \quad
\textbf{Yimeng Zhang}\textsuperscript{*} \quad
\textbf{Haoyu Gu} \quad
\textbf{Nuo Chen} \\
\textbf{Dong She} \quad
\textbf{Xianrong Yao} \quad
\textbf{Yang Gao}\textsuperscript{\textdagger} \quad
\textbf{Zhanpeng Jin}\textsuperscript{\textdagger} \\
School of Future Technology, South China University of Technology, Guangzhou, China \\
\texttt{\{202364870212, 202364870412, 202364820071, 202230310052\}@mail.scut.edu.cn}\\
\texttt{\{ftdshe,  ftxryao\}@mail.scut.edu.cn}\\
\texttt{\{gaoyang2025, zjin\}@scut.edu.cn}\\
\small
\textsuperscript{*}Equal contribution. \quad
\textsuperscript{\textdagger}Corresponding authors.
}%

\begin{document}
\maketitle
\begin{abstract}
Multimodal Emotion Recognition (MER) is critical for interpreting real-world interactions. While Multimodal Large Language Models (MLLM) have shown promise in MER, their internal decision-making mechanisms under modality conflict and missingness remain largely underexplored. In this paper, to systematically investigate these behaviors, we introduce EmoMM, a comprehensive benchmark featuring modality-aligned, conflict, and missing subsets. Through extensive evaluation, we uncover a Video Contribution Collapse (VCC) phenomenon, where MLLM marginalize video evidence due to high token redundancy and modality preferences. To address this, we propose Conflict-aware Head-level Attention Steering (CHASE), a lightweight mechanism that detects modality conflicts and performs inference-time attention steering, effectively mitigating decision bias without retraining the backbone. Experimental results demonstrate that CHASE consistently improves performance across various settings, significantly enhancing the reliability of MLLM in complex affective scenarios.
\end{abstract}

\section{Introduction}

\begin{figure}[t]
  \centering
  \includegraphics[width=\columnwidth]{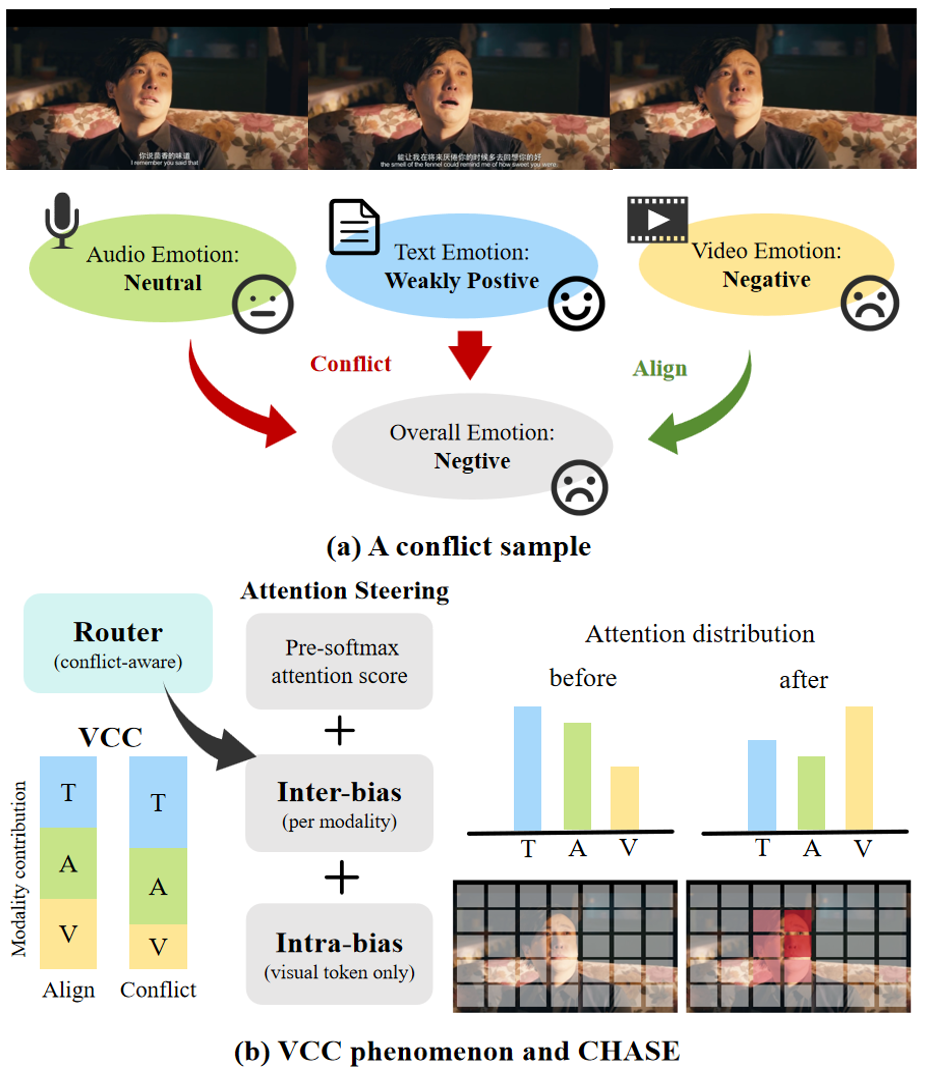}
  \caption{Overview of modality conflict in MER and our study. (a) A conflict sample. (b) The Video Contribution Collapse (VCC) phenomenon and our Conflict-aware Head-level Attention Steering (CHASE).}
  \label{fig:intro}
\end{figure}

Multimodal Emotion Recognition (MER)\citep{lai2023surveyInfoSci, yin2024mllmsurvey} aims to infer affect by jointly modeling text, visual behaviors, and acoustic prosody to better capture emotions in real-world interactions, supporting a wide range of applications such as education\citep{yan2025student_engagement_plosone, li2025learning_engagement_hci}, mental health\citep{sadeghi2024depression_llm_face, xu2025depression_voice_text}, and conversational systems\citep{kovacevic2024chatbot_mer_wild, fu2025multihm}. Real-world multimodal signals are often incomplete and sometimes contradictory, Figure ~\ref{fig:intro}(a) illustrates a conflict sample: the text conveys an affectionate sentiment and the vocal cues are subdued, while the speaker’s reddened eyes and trembling lips clearly reveal sadness and distress.

Recently, the multimodal large language model (MLLM) \citep{openai2024gpt4o, gemini2024gemini15, chu2024qwen2audio, xu2025qwen25omni, fu2025vita15, zhang2023videollama} paradigm has advanced rapidly, enabling unified cross-modal
reasoning over video, speech, and text, and showing strong potential on MER \citep{cheng2024emotionllama, lian2025affectgpt, fang2025catch, zhang2025mmeemotion}.
Recent work has begun to study robustness under conflict and missing-modality settings, by developing targeted evaluations and mitigation strategies that diagnose and reduce modality bias in MLLM\citep{han2025mosear, guo2024multimodalprompt, pipoli2025missrag}.
Modality missingness and modality conflict share a common abstraction as two complementary degradation axes of imperfect multimodal inputs: missingness removes evidence, while conflict introduces contradictory evidence. When modality evidence is imperfect, models should dynamically estimate modality trust and reweight evidence, which is why we study modality missingness and modality conflict together.
However, current work remains largely descriptive, while leaving unclear which internal evidence drives failures or when the decision process drifts from truly informative cues, so mechanistic analyses of these behaviors are still underexplored.

To systematically explain the decision-making mechanisms of MLLM in MER under modality conflict and missingness,
this paper investigates three research questions:\par\smallskip
\noindent\hspace*{1.5em}\textbullet\,\textbf{RQ1}: \textit{To what extent can current MLLM effectively integrate multimodal evidence to make decisions under conflict and missing conditions?}\\
\noindent\hspace*{1.5em}\textbullet\,\textbf{RQ2}: \textit{Does MLLM exhibit stable modality preference in MER?}\\
\noindent\hspace*{1.5em}\textbullet\,\textbf{RQ3}: \textit{How can such preference be mitigated to enable more reliable sentiment judgments under modality conflict and missingness?}\par\smallskip

To answer \textbf{RQ1} and \textbf{RQ2}, we first introduce \textbf{EmoMM}, a multimodal sentiment benchmark with modality aligned, conflict, and missing subsets, enabling a comprehensive evaluation of MLLMs' capability and internal mechanisms under modality conflict and missingness (see Sec. \ref{sec:Benchmark}). 
Then, we evaluate multiple MLLMs on \textbf{EmoMM} and uncover a \textbf{Video Contribution Collapse (VCC)} phenomenon (see Sec. \ref{sec:VCC}), where video contribution declines sharply in conflict, as illustrated in Figure ~\ref{fig:intro}(b) left. 
Combining metric-based evidence with attention analysis, we attribute VCC to the larger scale and higher redundancy of video tokens, together with the model's stable modality preference (see Sec. \ref{sec:Pre}). 

To answer \textbf{RQ3}, we propose \textbf{CHASE} (\textbf{C}onflict-aware \textbf{H}ead-level \textbf{A}ttention \textbf{S}teering), a lightweight, conflict-aware router that detects modality conflicts and then performs inference-time attention steering, ensuring that three modalities achieve an ideal attention distribution via inter-bias while putting more attention on critical visual tokens via intra-bias (see Sec.\ref{sec:method}), as sketched in Figure ~\ref{fig:intro}(b) right.
Experimental results show that this mechanism improves overall accuracy by 5.43\%. Notably, on challenging settings such as EmoMM-Conflict and EmoMM-Missing, it achieves accuracy gains of 5.16\% and 6.03\% respectively, substantially strengthening sentiment judgment under modality conflict and missingness.

Our main contributions are summarized as follows.\par\smallskip
\noindent\hspace*{1.5em}\textbullet\,\textbf{A Benchmark}: We introduce \textbf{EmoMM}, a multimodal sentiment benchmark with modality-aligned, conflict, and missing subsets for comprehensive evaluation of MLLM under conflict and missingness.\par
\noindent\hspace*{1.5em}\textbullet\,\textbf{Analysis and Findings}: We identify \textbf{VCC} and relate it to video-token redundancy, low evidence density, and MLLM's stable modality preference.\par
\noindent\hspace*{1.5em}\textbullet\,\textbf{Method}: We propose \textbf{CHASE}, a lightweight, conflict-aware router that detects modality conflicts and performs head-level, inference-time attention steering.\par

\section{Related Work}

\begin{figure*}[t]
  \centering
  \includegraphics[width=\textwidth]{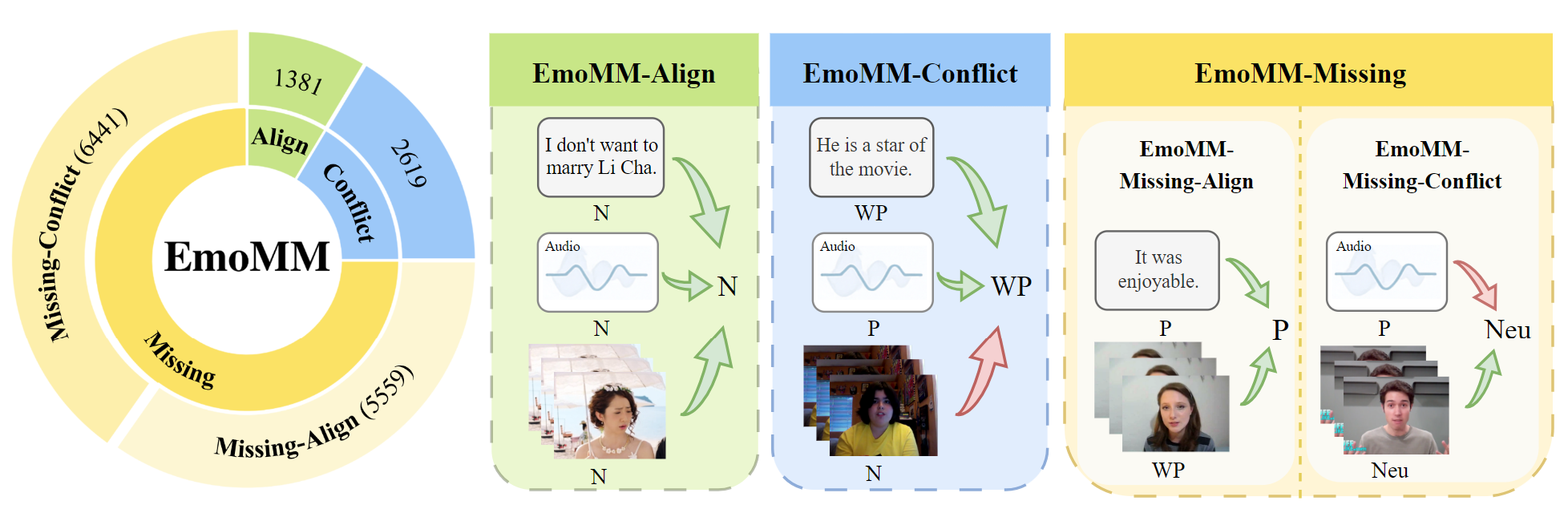}
  \caption{Overview of the EmoMM benchmark. Left: subset split of EmoMM into EmoMM-Align, EmoMM-Conflict, and EmoMM-Missing. Right: example instances illustrating the three subsets}
  \label{fig:subset}
\end{figure*}

\subsection{Multimodal Emotion Recognition}
Multimodal emotion recognition aims to fuse linguistic, acoustic, and visual signals to understand emotion. Early approaches predominantly relied on explicit fusion strategies, such as feature-level and tensor-level fusion~\citep{baltrusaitis2019mmlsurvey,lai2023surveyInfoSci,ngiam2011multimodal,zadeh2017tfn}. With the rapid advancement of deep learning, research has gradually shifted toward attention-based fusion paradigms, among which cross-modal interaction modeling with Multimodal Transformers has emerged as a mainstream solution~\citep{lai2023surveyInfoSci,tsai2019mult}. Most recently, this line of work has been further propelled by MLLM. General-purpose omni models enable unified audio--video--text interaction, such as GPT-5~\citep{openai2025gpt5}, Qwen2.5-Omni~\citep{xu2025qwen25omni}, VITA-1.5~\citep{fu2025vita15}, and MiniCPM-o 2.6~\citep{openbmb2025minicpmo26}. 

Beyond general-purpose capabilities in MER, the robustness and reliability of MLLM under modality conflict and missingness remains underexplored. MoSEAR introduces a conflict-aware benchmark (CA-MER), diagnoses systematic audio over-reliance in conflict cases, and proposes modality-specific experts with inference-time attention reallocation for rebalancing~\citep{han2025mosear}. For missingness, prompt and retrieval augmented approaches have been explored, and MissRAG further improves robustness via multimodal retrieval and modality-aware prompting~\citep{guo2024multimodalprompt,fan2024ramer,pipoli2025missrag}. To bridge the gap that a comprehensive understanding of how MLLM behaves under modality conflicts and missingness in MER remains limited, we propose \textbf{EmoMM}, which provides a unified benchmark for systematically evaluating and analyzing MLLM under modality conflict and missingness.

\subsection{Multimodal Emotion Recognition Datasets}
Progress in multimodal emotion recognition has been driven by a series of benchmark datasets. In English, mainstream datasets include IEMOCAP for dyadic dialogue emotion recognition~\citep{busso2008iemocap}; CMU-MOSI focusing on video-level subjective sentiment intensity~\citep{zadeh2016mosi} and its large-scale extension CMU-MOSEI, which covers a richer set of speakers and content~\citep{zadeh2018cmu-mosei}; and MELD, which targets multimodal emotions in multi-speaker conversations~\citep{poria2019meld}.
In Chinese, CH\mbox{-}SIMS innovatively provides independent single-modality annotations that enable the study of cross-modality inconsistency~\citep{yu2020chsims,thuiar2022chsimsV2page,liu2022chsimsv2}. Overall, mainstream datasets are predominantly English, cross-lingual settings remain limited. To fill the gap that there is a lack of benchmarks specifically designed for modality conflict and modality missingness, \textbf{EmoMM} is designed to systematically evaluate the robustness of MLLM.

\section{EmoMM Benchmark}\label{sec:Benchmark}

Multimodal affective signals are inherently heterogeneous, and real-world applications frequently encounter both cross-modal conflicts and modality missingness~\citep{baltrusaitis2019mmlsurvey,lai2023surveyInfoSci}.  However, most existing multimodal sentiment datasets only provide a unified multimodal annotation for each multimodal segment and mainly assume full-modality inputs~\citep{zadeh2016mosi,zadeh2018cmu-mosei,poria2019meld}, making it difficult to systematically investigate conflict and missingness scenarios. To fill this gap, we introduce the multimodal emotion recognition benchmark \textbf{EmoMM}, which provides both unimodal and multimodal emotion annotations, and supports evaluation under both full-modality and missing-modality inputs.

\subsection{Benchmark Construction}
EmoMM is built upon two classic datasets, the Chinese CH-SIMS v2.0 and the English CMU-MOSI~\citep{liu2022chsimsv2,zadeh2016mosi}. We sample 2,000 instances from each dataset to form a balanced Chinese–English bilingual benchmark. CH-SIMS v2.0 natively contains independent unimodal and multimodal sentiment labels~\citep{liu2022chsimsv2}; we adopt its five-class scheme (Negative (N), Weakly Negative (WN), Neutral (Neu), Weakly Positive (WP), Positive (P))~\citep{yu2020chsims}, and use stratified sampling to cover typical cases. The original CMU-MOSI dataset only provides multimodal sentiment labels~\citep{zadeh2016mosi}. To unify the annotation format, we map its labels into the same five-class scheme and select 2,000 samples with a relatively balanced distribution. Then, to obtain independent unimodal labels aligned with CH-SIMS v2.0, we recruit five trained annotators to label each sample under a unimodal-only visibility setting in the order of Text, Audio, and Video, to avoid cross-modal information contamination; each annotator assigns -1 (negative), 0 (neutral), or 1 (positive) to each modality, and we average the five scores and map them to the five-class labels.

\subsection{Subset Construction}
To evaluate MER under varying cross-modal conditions, we partition \textbf{EmoMM} into three subsets based on the polarity relationship between each modality and the multimodal label (GT): \textbf{EmoMM-Align} (all modalities’ polarities are consistent with the GT), \textbf{EmoMM-Conflict} (at least one modality’s polarity conflicts with the GT), and \textbf{EmoMM-Missing} (missing-modality samples constructed by controlled masking on full-modality samples). Furthermore, EmoMM-Missing is split into EmoMM-Missing-Align and EmoMM-Missing-Conflict according to whether the remaining observed modalities are all aligned with the GT or still contain conflicts. We visualize the detailed statistics and provide concrete examples for each subset in Figure~\ref{fig:subset}.

\subsection{Data Statistics and Analysis}
\label{sec:data_stats}
We summarize label distributions for GT, Text (T), Audio (A), and Visual (V) over the five ordered emotion classes,
and quantify pairwise consistency between any two sources
$i,j\in\{\mathrm{GT},\mathrm{T},\mathrm{A},\mathrm{V}\}$ using mapped integer labels $y \in \{0, \dots, 4\}$:
\begin{equation}
C_{ij}
= 1 - \frac{1}{4N}\sum_{n=1}^{N}\left|y_i^{(n)}-y_j^{(n)}\right|.
\label{eq:cij}
\end{equation}
where $N$ is the number of samples, and $y_i^{(n)}$ denotes the $n$-th label value in source $i$.
$C_{ij}=1$ implies perfect agreement, while values closer to $0$
indicate larger average ordinal deviations.

From the $C_{ij}$ matrix (see Appendix \ref{app:data_analysis}), we observe the strongest agreement between GT and text, suggesting that lexical cues often provide the most direct signal.
The weakest agreement appears between text and video, which aligns with prior findings that visual affect can be subtle and context-dependent~\citep{liu2022chsimsv2}.

\begin{figure*}[t]
\centering
\includegraphics[width=\textwidth]{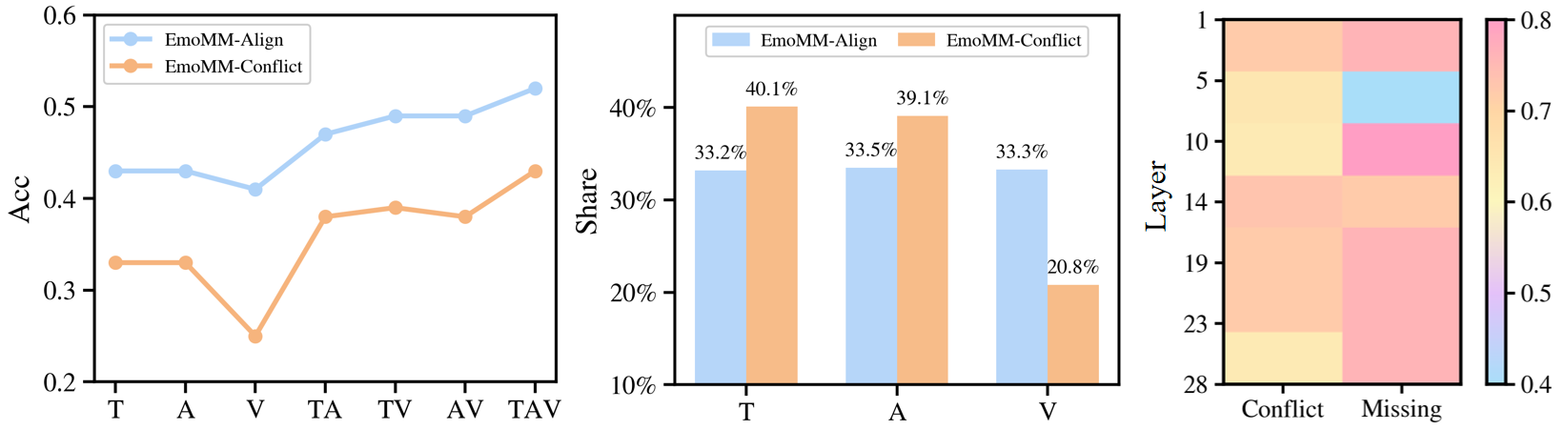}
\caption{Analyses of VCC. Left: Acc of different modality combinations. Medium: PSMV share under the Acc metric. Right: correlation between nMAS and GTAR.}
\label{fig:result}
\vspace{-0.3em}
\end{figure*}

\section{Understanding Decision-Making Mechanisms of MLLM under Modality Conflicts}

\subsection{Metrics}

\textbf{Performance Shapley Modality Value (PSMV).}
To measure the average marginal contribution of each modality across all possible modality \textbf{combinations},
we define PSMV inspired by the Shapley value~\citep{shapley1953value}.
Let $\mathcal{M}=\{T,A,V\}$ be the full modality set. For any modality subset $\mathcal{S}\subseteq\mathcal{M}$,
let $V(\mathcal{S})$ denote the model accuracy when only modalities in $\mathcal{S}$ are provided.
\begin{equation}
\begin{aligned}
\mathrm{PSMV}(m)
&= \sum_{m\in\mathcal{S}\subseteq \mathcal{M}}
w_{\mathcal{S}}\Big(V(\mathcal{S})-V(\mathcal{S}\setminus\{m\})\Big), \\
w_{\mathcal{S}}
&= \frac{(|\mathcal{S}|-1)!\,(|\mathcal{M}|-|\mathcal{S}|)!}{|\mathcal{M}|!}.
\end{aligned}
\label{eq:psmv}
\end{equation}
where $\mathcal{S}\setminus\{m\}$ is the modality subset obtained by removing modality $m$ from $\mathcal{S}$.

\textbf{Ground-Truth Aligned Rate (GTAR).}
To quantify how often the model prediction agrees with the ground-truth-aligned modality under conflict cases,
we define GTAR.
For sample $x$, let $\mathcal{M}_x\subseteq\mathcal{M}$ be the set of available modalities, $z_x^{m}$ be the polarity of modality $m$, $z_x^{GT}$ the ground-truth polarity, and $\hat{z}_x$ the model prediction.
We first introduce two binary variables to represent modality alignment ($a_x^{m}$) and prediction correctness ($c_x$):
\begin{equation}
\begin{aligned}
a_x^{m} &=
\begin{cases}
1, & \text{if } m\in\mathcal{M}_x \text{ and } z_x^{m}=z_x^{GT},\\
0, & \text{otherwise},
\end{cases}\\
c_x &=
\begin{cases}
1, & \text{if } \hat{z}_x=z_x^{GT},\\
0, & \text{otherwise}.
\end{cases}
\end{aligned}
\label{eq:gtar_defs}
\end{equation}

We focus on the target-conflict set $\mathcal{S}_{tc}$ including samples where exactly one available modality disagrees with the ground truth. Then, for each modality $m\in\mathcal{M}$, GTAR is defined as
\begin{equation}
\mathrm{GTAR}_m
=
\frac{
\sum\limits_{x\in\mathcal{S}_{tc}} a_x^{m}\,c_x
}{
\sum\limits_{x\in\mathcal{S}_{tc}} a_x^{m}
}.
\label{eq:gtar}
\end{equation}

\textbf{Normalized Mean Attention Score (nMAS).}
Let $K_m$ denote the token set of modality $m$.
For a decision query position $q$, we measure how much head $h$ at layer $\ell$
attends to modality $m$ by
\begin{equation}
\mathrm{nMAS}_{m}^{\ell,h}
= \frac{1}{|K_m|}\sum_{k \in K_m} A_{q,k}^{\ell,h},
\label{eq:mas}
\end{equation}
where $A_{q,k}^{\ell,h}$ is the (softmax-normalized) attention weight from the query token $q$
to key token $k$ in head $h$ of layer $\ell$ (More details about metrics in Appendix \ref{app:metrics_explanation}).

\subsection{Video Contribution Collapse (RQ1)}\label{sec:VCC}

To evaluate the multimodal integration ability of current MLLM under semantic conflict, we conduct a differential analysis on EmoMM-Align and EmoMM-Conflict subsets.
In Figure~\ref{fig:result} left, the results show that in EmoMM-Conflict, unimodal video performance degrades most significantly, with a drop of {16.0\%} clearly larger
than that of text ({10.4\%}) and audio ({9.8\%}), suggesting that video tends to become a weak signal or noise under conflicting contexts,
making it difficult for the model to stably extract visual evidence relevant to emotion recognition.
Further, in Figure~\ref{fig:result} medium, we observe that contributions are approximately balanced across
three modalities in EmoMM-Align, whereas EmoMM-Conflict exhibits a \textbf{Video Contribution Collapse (VCC)} phenomenon,
where the contribution distribution shifts markedly toward text and audio and the video contribution proportion drops significantly (from {33.3\%} to {20.8\%}).
Overall, the model calls complementary cues from the three modalities relatively evenly in consistent contexts,
but under conflicts it behaves more like adopting an evidence-availability strategy: prioritizing more stable evidence sources,
rather than actively mining explainable arbitration cues from video to correct other modalities.

\subsection{Modality Preference (RQ2)}\label{sec:Pre}

To investigate the potential causes of VCC, we further analyze GTAR on EmoMM-Conflict and  EmoMM-Missing-Conflict. The results show a consistent ordering of $T>A>V$ (see Figure\ref{fig:gtar_nmas}(a) in Appendix \ref{app:GTAR}), indicating a ‘‘cross-context stable’’ \textbf{modality preference}.
In conflict samples, video has the lowest GTAR, indicating that even when video is consistent with the overall emotion and carries correct emotional cues, the model still tends not to follow the video modality.

To explore attention-level causes, we introduce nMAS and analyze its correlation with GTAR.
In Figure~\ref{fig:result} right, the results show a significant positive correlation: when a modality receives higher attention density per token,
it is more likely to be adopted by the model when serving as the correct evidence source.
Therefore, improving GTAR is not about simply increasing a modality’s total attention, but rather about enhancing the
attention density and utilization efficiency of its effective evidence.

However, attention analysis shows that because the number of video tokens is much larger than that of text and audio,
even if video receives a non-trivial amount of total attention, its mean attention density per token (nMAS) remains far lower
than the other two modalities (see Figure
    \ref{fig:nmas_missing} and Figure \ref{fig:gtar_nmas}(b) in Appendix \ref{app:GTAR}). In addition, statistics on the spatial distribution of video attention in conflict samples show
that attention is often dispersed across many regions of frames, whereas the truly critical visual evidence for emotion recognition
is mostly concentrated in local regions such as facial micro-expressions, gaze, and head pose.
Many video tokens correspond to background or static areas; such redundant visual tokens consume the attention budget and reduce
effective evidence density, and are more likely to become noise sources under conflicting contexts.
In summary, we attribute the \textbf{key factors} of VCC to: \emph{the larger scale and higher redundancy of video tokens, which keeps video at a persistent disadvantage
in attention competition, together with the model's stable modality preference.}

To validate this hypothesis, we use MediaPipe to localize the person region and mask background tokens outside the person,
reducing the number of video tokens to the same order of magnitude as audio and improving video evidence density.
The results show that the marginal gain of video on
the final prediction turns from negative to positive, validating the hypothesis (see Appendix \ref{app:person_mask}).

\section{CHASE}
\label{sec:method}

To mitigate the performance degradation of MLLM under modality conflict and missingness (\textbf{RQ3}),
we propose \textbf{CHASE} (\textbf{C}onflict-aware \textbf{H}ead-level \textbf{A}ttention \textbf{S}teering).
CHASE inserts lightweight modules into a frozen MLLM and reallocates attention budgets at inference time
to reduce decision bias under modality conflicts.

\begin{figure}[t]
  \centering
  \includegraphics[width=\columnwidth]{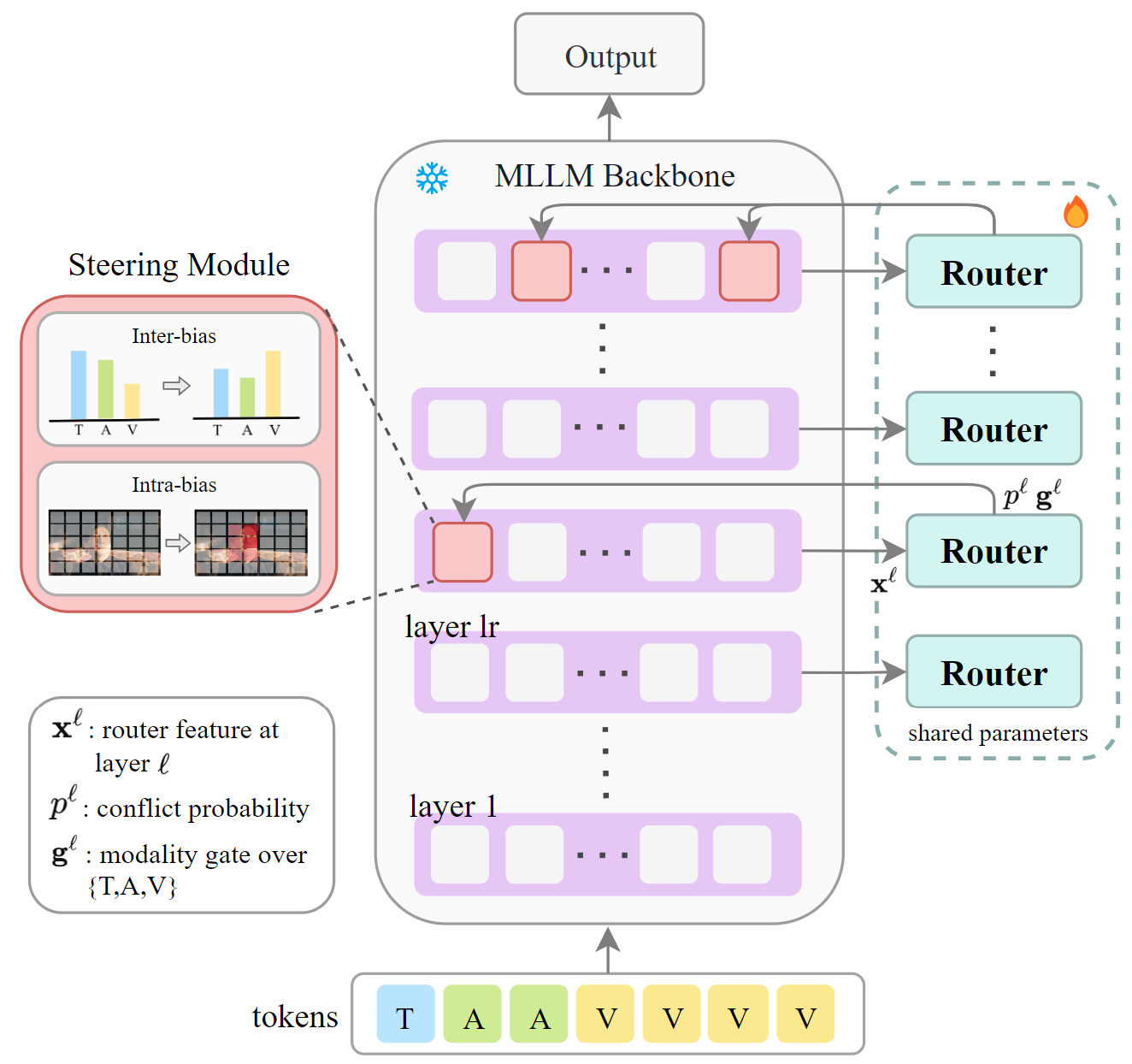}
  \caption{Illustration of the CHASE framework, where a router identifies conflicts and reallocates attention via inter- and intra-modality steering.}
  \label{fig:CHASE}
\end{figure}

\subsection{Conflict Detection Router}
\label{sec:router}

We deploy a lightweight Router on layers $\ell\in\{l_r,\dots,L\}$ and tie (share) its parameters across layers.
Let $\mathcal{M}=\{T,A,V\}$ be the full modality set and $\mathcal{M}_x\subseteq\mathcal{M}$ the modalities available for input $x$.

Based on the head-wise normalized Modality Attention Score (nMAS; Eq.~\eqref{eq:mas}),
we compute the per-head modality preference
$\boldsymbol{\pi}^{\ell,h}(x)$:
\begin{equation}
\boldsymbol{\pi}^{\ell,h}(x)
= \mathrm{softmax}\!\Big(
[\mathrm{nMAS}_{m}^{\ell,h}(x)]_{m \in \mathcal{M}}
+ \mathbf{E}(x)
\Big),
\label{eq:pi_head}
\end{equation}
where $\mathbf{E}(x)$ serves as a masking term to handle modality missingness, defined as $E_m(x)=0$ if $m\in\mathcal{M}_x$ and $E_m(x)=-\infty$ otherwise.
Then we derive the layer-level preference as
$\boldsymbol{\pi}^{\ell}(x)= \frac{1}{H}\sum_{h=1}^{H}\boldsymbol{\pi}^{\ell,h}(x)$.

The input to the Router, denoted as $\mathbf{x}^{\ell}$, is designed to capture both the semantic content and the model's current attention distribution.
At each layer $\ell$, the Router takes a feature vector $\mathbf{x}^{\ell} = \big[\mathbf{h}_{q}^{\ell} \,;\, \mathbf{h}_{T}^{\ell} \,;\, \mathbf{h}_{A}^{\ell} \,;\, \mathbf{h}_{V}^{\ell} \,;\, \boldsymbol{\pi}^{\ell}\big],
    \label{eq:router_in}$
where $\mathbf{h}_{q}^{\ell}$ is the query-token hidden state and $\mathbf{h}_{m}^{\ell}$ is a mean-pooled modality summary.
For missing modalities $m\notin\mathcal{M}_x$, we set $\mathbf{h}_{m}^{\ell}=\mathbf{0}$.

The Router consists of two lightweight projection heads (MLPs) that predict:

\vspace{0.2em}
\noindent $\bullet$ \textbf{Conflict Probability ($p^{\ell}$):}
A scalar estimate $p^{\ell}\in(0,1)$ indicating the likelihood that the current sample contains cross-modal conflicts at layer $\ell$:
\begin{equation}
p^{\ell} = \sigma(\mathbf{w}_p^\top \mathbf{x}^{\ell} + b_p),
\end{equation}
where $\sigma(\cdot)$ denotes the sigmoid activation function.

\vspace{0.2em}
\noindent $\bullet$ \textbf{Predicted Gate ($\mathbf{g}^{\ell}$):}
A probability distribution over available modalities that reflects the Router's estimated trustworthiness:
\begin{equation}
\mathbf{g}^{\ell}
= \mathrm{softmax}\!\Big( \mathbf{W}_g \mathbf{x}^{\ell} + \mathbf{b}_g + \mathbf{E} \Big),
\end{equation}
where $E_m=0$ if $m\in\mathcal{M}_x$ and $E_m=-\infty$ otherwise.

The conflict label $p^{\star}$ is set to 1 if any one unimodal label disagrees with the multimodal label (GT), and 0 otherwise. To teach the Router which modality to trust, we construct a soft target distribution $\mathbf{g}^\star(x)$. We define the correctness of modality $m$ as $r_m = \mathbb{I}(y_m = y_{gt})$. For all $m \in \mathcal{M}_x$, the target weight is computed by normalizing over available modalities: 

\begin{equation} 
g^\star_m = \frac{r_m + \epsilon}{\sum_{k \in \mathcal{M}_x} (r_k + \epsilon)}, 
\label{eq:target_gate} 
\end{equation} 
where $\epsilon > 0$ is a smoothing term to prevent zero probabilities. For missing modalities, $g^\star_m$ is set to 0. For instance, if audio conflicts with the GT while text and video align, $\mathbf{g}^{\star}$ is constructed to suppress audio and upweight the others.

We freeze the MLLM and optimize only the Router using a combined objective of binary cross-entropy for conflict detection
and KL-divergence for gate fitting over layers $\ell\in\{l_r,\dots,L\}$:
\begin{equation}
\mathcal{L}
= \sum_{\ell=l_r}^{L}
\Big(
\lambda_{1}\,D_{\mathrm{KL}}(\mathbf{g}^{\star}\,\|\,\mathbf{g}^{\ell})
+
\lambda_{2}\,\mathrm{BCE}(p^{\star},p^{\ell})
\Big),
\label{eq:router_loss}
\end{equation}
where $\lambda_{1}$ and $\lambda_{2}$ balance the two loss terms.

\subsection{Inference-time Attention Steering}
\label{sec:steering}

We intervene on the mid-to-late layers $\ell\in\{l_r,\dots,L\}$.
At inference time, we run the shared Router at each layer to obtain $(p^{\ell},\mathbf{g}^{\ell})$ and select layers by a conflict threshold: $\mathcal{L}=\{\ell\in\{l_r,\dots,L\}\mid p^{\ell}>\tau\}$, where $\tau$ is a hyperparameter.
For each selected layer $\ell\in\mathcal{L}$, we further select conflict-critical heads by
\begin{equation}
\mathcal{H}^{\ell}=\mathrm{TopK}_{h}\ \mathrm{JSD}\!\left(\boldsymbol{\pi}^{\ell,h}\,\|\,\mathbf{g}^{\ell}\right),
\label{eq:head_select}
\end{equation}
where $\mathrm{JSD}(\cdot\|\cdot)$ measures the discrepancy between the head preference and the Router gate
(see Appendix~\ref{app:loss_jsd}).

\vspace{0.2em}
\noindent $\bullet$ \textbf{Inter-modality Bias ($b^{\ell,h}_m$):}
For each selected head $(\ell,h)$ with $\ell\in\mathcal{L}$ and $h\in\mathcal{H}^{\ell}$, calculate the target attention distribution $\widehat{\boldsymbol{\pi}}^{\ell,h}$ by interpolating the original head preference $\boldsymbol{\pi}^{\ell,h}$ with the Router gate $\mathbf{g}^{\ell}$,
using $\eta p^{\ell}$ as the effective steering strength:
\begin{equation}
\widehat{\boldsymbol{\pi}}^{\ell,h}
=(1-\eta p^{\ell})\,\boldsymbol{\pi}^{\ell,h}+\eta p^{\ell}\,\mathbf{g}^{\ell},
\label{eq:quota}
\end{equation}
where $\eta\in[0,1]$ is a hyperparameter controlling the maximum steering magnitude.
We then convert the change in modality preference into an additive logit bias:
\begin{equation}
b^{\ell,h}_m=\log\frac{\widehat{\pi}^{\ell,h}_m+\epsilon_a}{\pi^{\ell,h}_m+\epsilon_a}.
\label{eq:inter_bias}
\end{equation}
where $\epsilon_a > 0$ is a smoothing term.

\vspace{0.2em}
\noindent $\bullet$ \textbf{Intra-modality Bias ($\delta_{qk}^{\ell,h}$):}
While the Router handles inter-modality conflict, we put more attention on critical visual tokens to address the problem that redundant visual tokens consume the attention budget. 

First, we compute the importance score $I_k$ for each token and normalize it to $\hat{{I}}_k$ (see Appendix \ref{sec:appendix_importance}).
We then define the significant visual token set ${S_v}$, containing the indices of the most informative video patches: ${S_v} = \left\{ k \in K_V \mid \hat{{I}}_k \geq \tau_P \right\}$, where $K_V$ is the token set, $\tau_P$ is the score threshold at the $P$-th percentile.

During the inference phase, our goal is for the attention mass allocated to the significant visual set $S_v$ to reach the target attention mass $\widehat{\psi}$. For a query $q$, we calculate the current attention mass $\psi{(q)} = \sum_{k \in {S_v}} A^{\ell,h}_{qk}$. If $\psi{(q)} < \widehat\psi$, we compute a compensatory bias $\delta_{qk}^{\ell,h}$ to bridge the gap:
\begin{equation}
\delta_{qk}^{\ell,h}=
\begin{cases}
\log\frac{\widehat{\psi}+\epsilon_b}{\psi(q)+\epsilon_b}, & k\in S_v \ \text{and}\ \psi(q) < \widehat{\psi},\\
0, & \text{otherwise.}
\end{cases}
\label{eq:intra_bias}
\end{equation}
where $ \widehat\psi$ is a hyperparameter, $\epsilon_b > 0$ is a smoothing term.

\vspace{0.2em}
\noindent $\bullet$ \textbf{Unified Attention Reallocation:}
Let $s_{qk}^{\ell,h}$ be the original pre-softmax attention score between query $q$ and key $k$ and we define the steered attention score $\tilde{s}_{qk}^{\ell,h}$ as:
\begin{equation}
    \tilde{s}_{qk}^{\ell,h} = s_{qk}^{\ell,h} + \gamma \cdot b^{\ell,h}_{m(k)} + \beta \cdot \delta_{qk}^{\ell,h}.
\end{equation}
where $m(k)$ denotes the modality of the \emph{key} token $k$, $\gamma$ and $\beta$ are hyperparameters controlling the strength of bias.

By using $\widetilde{s}^{(\ell,h)}_{ij}$ for the final softmax, CHASE ensures the decision is conditioned on reliable modalities (via $b^{\ell,h}_m$) while putting more attention on critical visual evidence (via $\delta_{qk}^{\ell,h}$).
\section{Experiments}
\label{sec:experiments}

In this section, we evaluate the effectiveness of \textbf{CHASE} on the proposed EmoMM benchmark.

\begin{table*}[t]
\centering
\resizebox{\textwidth}{!}{
\begin{tabular}{l|ccc|ccc|ccc|ccc}
\toprule
& \multicolumn{3}{c|}{\textbf{EmoMM (Overall)}} & \multicolumn{3}{c|}{\textbf{EmoMM-Align}} & \multicolumn{3}{c|}{\textbf{EmoMM-Conflict}} & \multicolumn{3}{c}{\textbf{EmoMM-Missing}} \\
\textbf{Model} & \textbf{Acc}$\uparrow$ & \textbf{F1}$\uparrow$ & \textbf{MAE}$\downarrow$ & \textbf{Acc}$\uparrow$ & \textbf{F1}$\uparrow$ & \textbf{MAE}$\downarrow$ & \textbf{Acc}$\uparrow$ & \textbf{F1}$\uparrow$ & \textbf{MAE}$\downarrow$ & \textbf{Acc}$\uparrow$ & \textbf{F1}$\uparrow$ & \textbf{MAE}$\downarrow$ \\
\midrule
\multicolumn{13}{l}{\textit{Open-source Baselines}} \\
Video-LLaMA2.1-7B-AV & 22.46 & 21.33 & 0.579 & 23.94 & 21.12 & 0.611 & 25.84 & 20.60 & 0.527 & 21.55 & 21.52 & 0.587 \\
MiniCPM-o 2.6-8B & 26.09 & 15.06 & 0.781 & 27.59 & 15.61 & 0.723 & 26.37 & 14.47 & 0.751 & 25.86 & 15.13 & 0.794 \\
Video-SALMONN 2-7B & 41.47 & 38.30 & 0.380 & 49.05 & 43.94 & 0.318 & 41.25 & 37.70 & 0.367 & 40.65 & 37.78 & 0.390 \\
\midrule
\multicolumn{13}{l}{\textit{Target Backbones \& Ours}} \\
Qwen2.5-Omni-7B & 44.53 & 36.24 & 0.384 & 52.09 & 44.28 & 0.270 & 46.47 & 39.76 & 0.356 & 43.27 & 34.55 & 0.403 \\
\rowcolor{gray!10} \textbf{+ CHASE (Ours)} & \textbf{49.96}  & 39.87 & 0.326 & 53.23 & 39.61 & \textbf{0.256} & 51.32 & 46.47 & 0.302 & \textbf{49.30} & 38.42 & 0.339 \\
\midrule
VITA-1.5-8B & 44.48 & 43.83 & 0.327 & 53.05 & 53.47 & 0.277 & 47.80 & 47.18 & 0.319 & 42.80 & 41.99 & 0.335 \\
\rowcolor{gray!10} \textbf{+ CHASE (Ours)} & 49.56 & \textbf{48.44} & \textbf{0.291} & \textbf{53.82} & \textbf{54.23} & 0.274 & \textbf{52.96} & \textbf{51.53} & \textbf{0.292} & 48.34 & \textbf{47.12} & \textbf{0.293} \\
\midrule
\multicolumn{13}{l}{\textit{Closed-source API}} \\
Gemini 3 Pro & 56.41 & 53.58 & 0.236 & 59.85 & 59.07 & 0.191 & 56.23 & 51.16 & 0.236 & 56.06 & 53.47 & 0.241 \\
\bottomrule
\end{tabular}
}
\caption{Main results on EmoMM and its subsets. CHASE consistently improves performance across both backbones, with the most significant gains observed in Conflict and Missing subsets.}
\label{tab:main_results}
\end{table*}

\subsection{Experimental Setup}

\paragraph{Datasets and Metrics.}
We conduct all evaluations on the \textbf{EmoMM} benchmark, covering 3 subsets: EmoMM-Align, EmoMM-Conflict, and EmoMM-Missing. 
We adopt a speaker-independent split strategy, partitioning the dataset into training, validation, and test sets with a ratio of 7:1:2. Furthermore, we use stratified sampling to ensure consistent label and category distributions across splits.
We report Accuracy (Acc), Macro F1-score (F1), and Mean Absolute Error (MAE) between predicted and ground-truth discrete emotion scores.

\paragraph{Model Configurations.}
We use official pre-trained weights for all baselines, including Video-LLaMA2.1-7B-AV~\citep{cheng2024videollama2}, MiniCPM-o 2.6-8B~\citep{openbmb2025minicpmo26} and Video-SALMONN 2-7B~\citep{sun2024videosalmonn}. We evaluate Gemini 3 Pro~\citep{gemini3pro2025} as a closed-source API baseline via the Gemini API.
To test universality, we apply CHASE to two target MLLM backbones:
Qwen2.5-Omni-7B~\citep{xu2025qwen25omni} and VITA-1.5-8B~\citep{fu2024vita}.

\paragraph{Implementation Details.}
We integrate the shared Router starting from layer $l_r{=}13$, implemented as a lightweight 2-layer MLP with a hidden dimension of 128, while the MLLM backbone remains entirely frozen.
For training, we employ the AdamW optimizer with a learning rate of 1e-4, a weight decay of 0.01, and a batch size of 32 for 10 epochs. The loss balancing coefficients are set to $\lambda_1{=}1.0$ and $\lambda_2{=}1.0$.
During inference, we apply CHASE with a fixed set of steering hyperparameters: the conflict detection threshold $\tau$ is set to $0.7$, and the maximum inter-modality steering coefficient $\eta$ is set to $0.6$. We intervene on the top-$K{=}4$ attention heads. The inter-modality steering strength $\gamma$ is set to $1.4$, and the intra-modality visual enhancement strength $\beta$ is set to $0.5$.
All experiments are conducted on NVIDIA A100 GPUs.

\begin{table}[ht]
\centering
\resizebox{0.9\columnwidth}{!}{
\begin{tabular}{l|ccc}
\toprule
\textbf{Variant (Qwen2.5-Omni)} & \textbf{Acc}$\uparrow$ & \textbf{F1}$\uparrow$ & \textbf{MAE}$\downarrow$ \\
\midrule
\rowcolor{gray!10} \textbf{Full CHASE} & \textbf{51.32} & \textbf{46.47} & \textbf{0.302} \\
\midrule
w/o Router (Random Steering) & 45.63 & 41.13 & 0.362 \\
w/o Inter-modality Bias ($b^{\ell,h}_m$) & 48.10 & 43.15 & 0.320 \\
w/o Intra-modality Bias ($\delta^{\ell,h}_{qk}$) & 50.15 & 46.50 & 0.298 \\
\midrule
Original Baseline & 46.47 & 39.76 & 0.356 \\
\bottomrule
\end{tabular}
}
\caption{Ablation study on the EmoMM-Conflict subset. We measure the impact of removing key components from the full CHASE framework.}
\label{tab:ablation}
\end{table}

\subsection{Main Results}
\label{sec:main_results}

\noindent\textbf{Robustness under Modality Conflict.}
CHASE enhances the robustness of MLLMs when facing contradictory semantic cues. As shown in Table~\ref{tab:main_results}, the improvements are particularly pronounced for Qwen2.5-Omni, which achieves an accuracy gain of 4.85\% on the EmoMM-Conflict subset. VITA-1.5 also benefits from our approach with a 5.16\% improvement. This confirms that our method effectively detects semantic conflicts and steers attention toward reliable modalities, thereby mitigating decision bias.

\noindent\textbf{Overall Performance and Missingness.}
On the aggregated test set (Overall), Qwen2.5-Omni equipped with CHASE achieves an accuracy of 49.96\%, representing a 5.43\% absolute improvement over the baseline. Moreover, Qwen2.5-Omni exhibits recoverability in the EmoMM-Missing subset, where its accuracy rises by 6.03\%. This confirms that the Router's gating mechanism successfully suppresses noise from missing modalities. Notably, on the EmoMM-Align subset, CHASE slightly improves performance (1.14\%), validating that our conflict detection mechanism has a low false-positive rate and preserves the model's reasoning capabilities when modalities are consistent.

\subsection{Ablation Study}
\label{sec:ablation}

To validate the effectiveness of each module, we conduct an ablation study on the EmoMM-Conflict subset using Qwen2.5-Omni (Table~\ref{tab:ablation}).

\noindent \textbf{Necessity of the Router:} Replacing our learned Router with random steering drops performance to 45.63\%. This confirms that precise conflict detection is essential, whereas blind intervention disrupts semantic processing.

\noindent \textbf{Impact of Inter-modality Steering:} Removing modality re-weighting ($b^{\ell,h}_m$) causes a 3.22\% decline. This indicates that simply enhancing visual tokens is insufficient without explicitly correcting the head-level modality hierarchy.

\noindent \textbf{Benefit of Intra-modality Bias:} Removing token densification ($\delta^{\ell,h}_{qk}$) decreases accuracy by 1.17\%. This module complements steering by filtering background noise to focus on emotion-relevant regions.

\section{Conclusion}
In this paper, we systematically investigated the decision-making mechanisms of Multimodal Large Language Models (MLLM) under realistic conditions of modality conflict and missingness. We introduced \textbf{EmoMM}, a comprehensive benchmark that facilitated the discovery of the Video Contribution Collapse (VCC) phenomenon, where models marginalize visual evidence due to token redundancy and stable modality preferences. To address this, we proposed \textbf{CHASE}, a lightweight mechanism that detects conflicts and steers head-level attention at inference time to rebalance multimodal evidence. Experimental results demonstrate that CHASE effectively mitigates decision bias and significantly enhances the robustness and reliability of MLLM in complex affective scenarios without the need for backbone retraining.

\section*{Limitations}
Despite the promising results, our study has several limitations. First, CHASE is evaluated mainly on multimodal emotion recognition, and its effectiveness on other multimodal reasoning tasks with different conflict patterns remains to be validated; moreover, as an inference-time intervention, it depends on the frozen backbone’s representational quality and cannot fundamentally repair cases where the base model fails to capture subtle cues. Second, CHASE requires white-box access to internal model signals (layer-wise hidden states and attention-related statistics) to estimate preferences and apply steering, which prevents direct deployment on purely black-box, API-only MLLMs. Third, the Router is trained with supervision derived from unimodal labels aligned with multimodal ground truth; extending CHASE to datasets with only multimodal labels is left for future work, potentially via pseudo-labels from modality-specific teachers, weak supervision from self-consistency under modality dropout, or confidence-based distillation signals.

\section*{Acknowledgments}
This work was supported in part by the National Natural Science Foundation of China (NSFC) under Grants W2532047 and 62302168; the Guangdong Provincial Department of Science and Technology under Grant 2023CX10X070; the Guangdong Provincial Key Laboratory of Human Digital Twin under Grant 2022B1212010004; the Guangzhou Basic Research Program under Grant SL2023A04J00930; and the Shenzhen Holdfound Foundation Endowed Professorship. We thank the annotators for their efforts in data annotation, and the reviewers and area chairs for their helpful suggestions.

\bibliography{custom}

\begin{thebibliography}{40}
\providecommand{\natexlab}[1]{#1}

\bibitem[{Baltru{\v{s}}aitis et~al.(2019)Baltru{\v{s}}aitis, Ahuja, and Morency}]{baltrusaitis2019mmlsurvey}
Tadas Baltru{\v{s}}aitis, Chaitanya Ahuja, and Louis-Philippe Morency. 2019.
\newblock \href {https://doi.org/10.1109/TPAMI.2018.2798607} {Multimodal machine learning: A survey and taxonomy}.
\newblock \emph{IEEE Transactions on Pattern Analysis and Machine Intelligence}, 41(2):423--443.

\bibitem[{Busso et~al.(2008)Busso, Bulut, Lee, Kazemzadeh, Mower, Kim, Chang, Lee, and Narayanan}]{busso2008iemocap}
Carlos Busso, Murtaza Bulut, Chi-Chun Lee, Abe Kazemzadeh, Emily Mower, Samuel Kim, Jeannette~N. Chang, Sungbok Lee, and Shrikanth~S. Narayanan. 2008.
\newblock \href {https://doi.org/10.1007/s10579-008-9076-6} {{IEMOCAP}: Interactive emotional dyadic motion capture database}.
\newblock \emph{Language Resources and Evaluation}, 42(4):335--359.

\bibitem[{Cheng et~al.(2024{\natexlab{a}})Cheng, Cheng, He, Sun, Wang, Lin, Lian, Peng, and Hauptmann}]{cheng2024emotionllama}
Zebang Cheng, Zhi-Qi Cheng, Jun-Yan He, Jingdong Sun, Kai Wang, Yuxiang Lin, Zheng Lian, Xiaojiang Peng, and Alexander~G. Hauptmann. 2024{\natexlab{a}}.
\newblock \href {https://proceedings.neurips.cc/paper_files/paper/2024/hash/c7f43ada17acc234f568dc66da527418-Abstract-Conference.html} {{Emotion-LLaMA}: Multimodal emotion recognition and reasoning with instruction tuning}.
\newblock \emph{Advances in Neural Information Processing Systems}, 37:110805--110853.

\bibitem[{Cheng et~al.(2024{\natexlab{b}})Cheng, Leng, Zhang, Xin, Li, Chen, Zhu, Zhang, Luo, Zhao, and Bing}]{cheng2024videollama2}
Zesen Cheng, Sicong Leng, Hang Zhang, Yifei Xin, Xin Li, Guanzheng Chen, Yongxin Zhu, Wenqi Zhang, Ziyang Luo, Deli Zhao, and Lidong Bing. 2024{\natexlab{b}}.
\newblock \href {https://arxiv.org/abs/2406.07476} {{VideoLLaMA 2}: Advancing spatial-temporal modeling and audio understanding in video-{LLM}s}.
\newblock \emph{arXiv preprint arXiv:2406.07476}.

\bibitem[{Chu et~al.(2024)Chu, Xu, Yang, Wei, Wei, Guo, Leng, Lv, He, Lin, Zhou, and Zhou}]{chu2024qwen2audio}
Yunfei Chu, Jin Xu, Qian Yang, Haojie Wei, Xipin Wei, Zhifang Guo, Yichong Leng, Yuanjun Lv, Jinzheng He, Junyang Lin, Chang Zhou, and Jingren Zhou. 2024.
\newblock \href {https://arxiv.org/abs/2407.10759} {{Qwen2-Audio} technical report}.
\newblock \emph{arXiv preprint arXiv:2407.10759}.

\bibitem[{Fan et~al.(2024)Fan, Yuan, Zuo, Liu, and Gao}]{fan2024ramer}
Qi~Fan, Hongyu Yuan, Haolin Zuo, Rui Liu, and Guanglai Gao. 2024.
\newblock \href {https://arxiv.org/abs/2410.02804} {Leveraging retrieval augment approach for multimodal emotion recognition under missing modalities}.
\newblock \emph{arXiv preprint arXiv:2410.02804}.

\bibitem[{Fang et~al.(2025)Fang, Liang, Huang, Li, Su, and Ye}]{fang2025catch}
Yiyang Fang, Jian Liang, Wenke Huang, He~Li, Kehua Su, and Mang Ye. 2025.
\newblock \href {https://proceedings.mlr.press/v267/fang25h.html} {Catch your emotion: Sharpening emotion perception in multimodal large language models}.
\newblock In \emph{Proceedings of the 42nd International Conference on Machine Learning (ICML)}, pages 16029--16039. PMLR.

\bibitem[{Fu et~al.(2024)Fu, Lin, Long, Shen, Dai, Zhao, Zhang, Dong, Li, Wang, Cao, Yin, Ma, Zheng, Ji, Wu, He, Shan, and Sun}]{fu2024vita}
Chaoyou Fu, Haojia Lin, Zuwei Long, Yunhang Shen, Yuhang Dai, Meng Zhao, Yi-Fan Zhang, Shaoqi Dong, Yangze Li, Xiong Wang, Haoyu Cao, Di~Yin, Long Ma, Xiawu Zheng, Rongrong Ji, Yunsheng Wu, Ran He, Caifeng Shan, and Xing Sun. 2024.
\newblock \href {https://arxiv.org/abs/2408.05211} {{VITA}: Towards open-source interactive omni multimodal {LLM}}.
\newblock \emph{arXiv preprint arXiv:2408.05211}.
\newblock Original version.

\bibitem[{Fu et~al.(2025{\natexlab{a}})Fu, Lin, Wang, Zhang, Shen, Liu, Cao, Long, Gao, Li, Ma, Zheng, Ji, Sun, Shan, and He}]{fu2025vita15}
Chaoyou Fu, Haojia Lin, Xiong Wang, Yi-Fan Zhang, Yunhang Shen, Xiaoyu Liu, Haoyu Cao, Zuwei Long, Heting Gao, Ke~Li, Long Ma, Xiawu Zheng, Rongrong Ji, Xing Sun, Caifeng Shan, and Ran He. 2025{\natexlab{a}}.
\newblock \href {https://arxiv.org/abs/2501.01957} {{VITA-1.5}: Towards {GPT-4o} level real-time vision and speech interaction}.
\newblock \emph{arXiv preprint arXiv:2501.01957}.

\bibitem[{Fu et~al.(2025{\natexlab{b}})Fu, Liu, Song, Zhang, and Liao}]{fu2025multihm}
Yao Fu, Qiong Liu, Qing Song, Pengzhou Zhang, and Gongdong Liao. 2025{\natexlab{b}}.
\newblock \href {https://doi.org/10.3390/app15084509} {{Multi-HM}: A {Chinese} multimodal dataset and fusion framework for emotion recognition in human--machine dialogue systems}.
\newblock \emph{Applied Sciences}, 15(8):4509.

\bibitem[{{Gemini Team}(2024)}]{gemini2024gemini15}
{Gemini Team}. 2024.
\newblock \href {https://arxiv.org/abs/2403.05530} {Gemini 1.5: Unlocking multimodal understanding across millions of tokens of context}.
\newblock \emph{arXiv preprint arXiv:2403.05530}.

\bibitem[{{Google DeepMind}(2025)}]{gemini3pro2025}
{Google DeepMind}. 2025.
\newblock \href {https://storage.googleapis.com/deepmind-media/Model-Cards/Gemini-3-Pro-Model-Card.pdf} {Gemini 3 pro model card}.
\newblock Technical report, Google DeepMind.

\bibitem[{Guo et~al.(2024)Guo, Jin, and Zhao}]{guo2024multimodalprompt}
Zirun Guo, Tao Jin, and Zhou Zhao. 2024.
\newblock \href {https://doi.org/10.18653/v1/2024.acl-long.94} {Multimodal prompt learning with missing modalities for sentiment analysis and emotion recognition}.
\newblock In \emph{Proceedings of the 62nd Annual Meeting of the Association for Computational Linguistics (Volume 1: Long Papers)}, pages 1714--1730. Association for Computational Linguistics.

\bibitem[{Han et~al.(2025)Han, Zhu, Xu, Song, and Yang}]{han2025mosear}
Zhiyuan Han, Beier Zhu, Yanlong Xu, Peipei Song, and Xun Yang. 2025.
\newblock \href {https://doi.org/10.1145/3746027.3754856} {Benchmarking and bridging emotion conflicts for multimodal emotion reasoning}.
\newblock In \emph{Proceedings of the 33rd ACM International Conference on Multimedia (MM '25)}. ACM.
\newblock ArXiv:2508.01181.

\bibitem[{Kova{\v{c}}evi{\'c} et~al.(2024)Kova{\v{c}}evi{\'c}, Holz, Gross, and Wampfler}]{kovacevic2024chatbot_mer_wild}
Nikola Kova{\v{c}}evi{\'c}, Christian Holz, Markus Gross, and Rafael Wampfler. 2024.
\newblock \href {https://doi.org/10.1145/3678957.3685759} {On multimodal emotion recognition for human-chatbot interaction in the wild}.
\newblock In \emph{Proceedings of the 26th International Conference on Multimodal Interaction (ICMI '24)}, pages 12--21. ACM.

\bibitem[{Lai et~al.(2023)Lai, Hu, Xu, Ren, and Liu}]{lai2023surveyInfoSci}
Songning Lai, Xifeng Hu, Haoxuan Xu, Zhaoxia Ren, and Zhi Liu. 2023.
\newblock \href {https://doi.org/10.1016/j.displa.2023.102563} {Multimodal sentiment analysis: A survey}.
\newblock \emph{Information Sciences}.

\bibitem[{Li et~al.(2025)Li, Weng, Li, and Zhang}]{li2025learning_engagement_hci}
Chuangqi Li, Xinying Weng, Yifan Li, and Tianxin Zhang. 2025.
\newblock \href {https://doi.org/10.1080/10447318.2024.2338616} {Multimodal learning engagement assessment system: An innovative approach to optimizing learning engagement}.
\newblock \emph{International Journal of Human--Computer Interaction}, 41(5):3474--3490.

\bibitem[{Lian et~al.(2025)Lian, Chen, Chen, Sun, Sun, Ren, Cheng, Liu, Liu, Peng, Yi, and Tao}]{lian2025affectgpt}
Zheng Lian, Haoyu Chen, Lan Chen, Haiyang Sun, Licai Sun, Yong Ren, Zebang Cheng, Bin Liu, Rui Liu, Xiaojiang Peng, Jiangyan Yi, and Jianhua Tao. 2025.
\newblock \href {https://proceedings.mlr.press/v267/lian25a.html} {{AffectGPT}: A new dataset, model, and benchmark for emotion understanding with multimodal large language models}.
\newblock In \emph{Proceedings of the 42nd International Conference on Machine Learning (ICML)}, pages 36993--37014. PMLR.

\bibitem[{Liu et~al.(2022)Liu, Yuan, Mao, Liang, Yang, Qiu, Cheng, Li, Xu, and Gao}]{liu2022chsimsv2}
Yihe Liu, Ziqi Yuan, Huisheng Mao, Zhiyun Liang, Wanqiuyue Yang, Yuanzhe Qiu, Tie Cheng, Xiaoteng Li, Hua Xu, and Kai Gao. 2022.
\newblock \href {https://doi.org/10.1145/3536221.3556630} {Make acoustic and visual cues matter: {CH-SIMS} v2.0 dataset and {AV-Mixup} consistent module}.
\newblock In \emph{Proceedings of the 2022 International Conference on Multimodal Interaction}, pages 224--232. ACM.

\bibitem[{Ngiam et~al.(2011)Ngiam, Khosla, Kim, Nam, Lee, and Ng}]{ngiam2011multimodal}
Jiquan Ngiam, Aditya Khosla, Minjae Kim, Juhan Nam, Honglak Lee, and Andrew~Y. Ng. 2011.
\newblock \href {https://dl.acm.org/doi/10.5555/3104482.3104569} {Multimodal deep learning}.
\newblock In \emph{Proceedings of the 28th International Conference on Machine Learning (ICML)}, pages 689--696.

\bibitem[{{OpenAI}(2024)}]{openai2024gpt4o}
{OpenAI}. 2024.
\newblock \href {https://arxiv.org/abs/2410.21276} {{GPT-4o} system card}.
\newblock \emph{arXiv preprint arXiv:2410.21276}.

\bibitem[{{OpenAI}(2025)}]{openai2025gpt5}
{OpenAI}. 2025.
\newblock \href {https://openai.com/index/introducing-gpt-5/} {Introducing {GPT-5}}.
\newblock Product announcement.

\bibitem[{{OpenBMB}(2025)}]{openbmb2025minicpmo26}
{OpenBMB}. 2025.
\newblock \href {https://huggingface.co/openbmb/MiniCPM-o-2_6} {{MiniCPM-o 2.6}: A {GPT-4o}-level {MLLM} for vision, speech and live streaming}.
\newblock Technical blog and model card.

\bibitem[{Pipoli et~al.(2025)Pipoli, Saporita, Bolelli, Cornia, Baraldi, Grana, Cucchiara, and Ficarra}]{pipoli2025missrag}
Vittorio Pipoli, Alessia Saporita, Federico Bolelli, Marcella Cornia, Lorenzo Baraldi, Costantino Grana, Rita Cucchiara, and Elisa Ficarra. 2025.
\newblock \href {https://openaccess.thecvf.com/content/ICCV2025/html/Pipoli_MissRAG_Addressing_the_Missing_Modality_Challenge_in_Multimodal_Large_Language_ICCV_2025_paper.html} {{MissRAG}: Addressing the missing modality challenge in multimodal large language models}.
\newblock In \emph{Proceedings of the IEEE/CVF International Conference on Computer Vision (ICCV)}.

\bibitem[{Poria et~al.(2019)Poria, Hazarika, Majumder, Naik, Cambria, and Mihalcea}]{poria2019meld}
Soujanya Poria, Devamanyu Hazarika, Navonil Majumder, Gautam Naik, Erik Cambria, and Rada Mihalcea. 2019.
\newblock \href {https://doi.org/10.18653/v1/P19-1050} {{MELD}: A multimodal multi-party dataset for emotion recognition in conversations}.
\newblock In \emph{Proceedings of the 57th Annual Meeting of the Association for Computational Linguistics}, pages 527--536. Association for Computational Linguistics.

\bibitem[{Sadeghi et~al.(2024)Sadeghi, Richer, Egger, Schindler-Gmelch, Rupp, Rahimi, Berking, and Eskofier}]{sadeghi2024depression_llm_face}
Misha Sadeghi, Robert Richer, Bernhard Egger, Lena Schindler-Gmelch, Lydia~Helene Rupp, Farnaz Rahimi, Matthias Berking, and Bjoern~M. Eskofier. 2024.
\newblock \href {https://doi.org/10.1038/s44184-024-00112-8} {Harnessing multimodal approaches for depression detection using large language models and facial expressions}.
\newblock \emph{npj Mental Health Research}, 3:66.

\bibitem[{Shapley(1953)}]{shapley1953value}
Lloyd~S. Shapley. 1953.
\newblock \href {https://doi.org/10.1515/9781400881970-018} {A value for n-person games}.
\newblock In Harold~W. Kuhn and Albert~W. Tucker, editors, \emph{Contributions to the Theory of Games II (Annals of Mathematics Studies, Vol. 28)}, pages 307--317. Princeton University Press.

\bibitem[{Sun et~al.(2024)Sun, Yu, Tang, Chen, Tan, Li, Lu, Ma, Wang, and Zhang}]{sun2024videosalmonn}
Guangzhi Sun, Wenyi Yu, Changli Tang, Xianzhao Chen, Tian Tan, Wei Li, Lu~Lu, Zejun Ma, Yuxuan Wang, and Chao Zhang. 2024.
\newblock \href {https://arxiv.org/abs/2406.15704} {video-salmonn: Speech-enhanced audio-visual large language models}.
\newblock \emph{arXiv preprint arXiv:2406.15704}.

\bibitem[{{THUIAR}(2022)}]{thuiar2022chsimsV2page}
{THUIAR}. 2022.
\newblock \href {https://github.com/thuiar/ch-sims-v2} {{CH-SIMS} v2.0 project page}.
\newblock Project page.

\bibitem[{Tsai et~al.(2019)Tsai, Bai, Liang, Kolter, Morency, and Salakhutdinov}]{tsai2019mult}
Yao-Hung~Hubert Tsai, Shaojie Bai, Paul~Pu Liang, J.~Zico Kolter, Louis-Philippe Morency, and Ruslan Salakhutdinov. 2019.
\newblock \href {https://doi.org/10.18653/v1/P19-1656} {Multimodal transformer for unaligned multimodal language sequences}.
\newblock In \emph{Proceedings of the 57th Annual Meeting of the Association for Computational Linguistics}, pages 6558--6569. Association for Computational Linguistics.

\bibitem[{Xu et~al.(2025{\natexlab{a}})Xu, Guo, He, Hu, He, Bai, Chen, Wang, Fan, Dang, Zhang, Wang, Chu, and Lin}]{xu2025qwen25omni}
Jin Xu, Zhifang Guo, Jinzheng He, Hangrui Hu, Ting He, Shuai Bai, Keqin Chen, Jialin Wang, Yang Fan, Kai Dang, Bin Zhang, Xiong Wang, Yunfei Chu, and Junyang Lin. 2025{\natexlab{a}}.
\newblock \href {https://arxiv.org/abs/2503.20215} {{Qwen2.5-Omni} technical report}.
\newblock \emph{arXiv preprint arXiv:2503.20215}.

\bibitem[{Xu et~al.(2025{\natexlab{b}})Xu, Gao, Wang, Zhang, Zhang, Wang, and Shu}]{xu2025depression_voice_text}
Zhenrong Xu, Yuan Gao, Fang Wang, Longqian Zhang, Li~Zhang, Junke Wang, and Jie Shu. 2025{\natexlab{b}}.
\newblock \href {https://doi.org/10.1038/s41598-025-03524-4} {Depression detection methods based on multimodal fusion of voice and text}.
\newblock \emph{Scientific Reports}, 15:21907.

\bibitem[{Yan et~al.(2025)Yan, Wu, and Wang}]{yan2025student_engagement_plosone}
Lijuan Yan, Xiaotao Wu, and Yi~Wang. 2025.
\newblock \href {https://doi.org/10.1371/journal.pone.0325377} {Student engagement assessment using multimodal deep learning}.
\newblock \emph{PLOS ONE}, 20(6):e0325377.

\bibitem[{Yin et~al.(2024)Yin, Fu, Zhao, Li, Sun, Xu, and Chen}]{yin2024mllmsurvey}
Shukang Yin, Chaoyou Fu, Sirui Zhao, Ke~Li, Xing Sun, Tong Xu, and Enhong Chen. 2024.
\newblock \href {https://doi.org/10.1093/nsr/nwae403} {A survey on multimodal large language models}.
\newblock \emph{National Science Review}, 11(12):nwae403.

\bibitem[{Yu et~al.(2020)Yu, Xu, Meng, Zhu, Ma, Wu, Zou, and Yang}]{yu2020chsims}
Wenmeng Yu, Hua Xu, Fanyang Meng, Yilin Zhu, Yixiao Ma, Jiele Wu, Jiyun Zou, and Kaicheng Yang. 2020.
\newblock \href {https://doi.org/10.18653/v1/2020.acl-main.343} {{CH-SIMS}: A {Chinese} multimodal sentiment analysis dataset with fine-grained annotation of modality}.
\newblock In \emph{Proceedings of the 58th Annual Meeting of the Association for Computational Linguistics}, pages 3718--3727. Association for Computational Linguistics.

\bibitem[{Zadeh et~al.(2017)Zadeh, Chen, Poria, Cambria, and Morency}]{zadeh2017tfn}
Amir Zadeh, Minghai Chen, Soujanya Poria, Erik Cambria, and Louis-Philippe Morency. 2017.
\newblock \href {https://doi.org/10.18653/v1/D17-1115} {Tensor fusion network for multimodal sentiment analysis}.
\newblock In \emph{Proceedings of the 2017 Conference on Empirical Methods in Natural Language Processing}, pages 1103--1114. Association for Computational Linguistics.

\bibitem[{Zadeh et~al.(2016)Zadeh, Zellers, Pincus, and Morency}]{zadeh2016mosi}
Amir Zadeh, Rowan Zellers, Eli Pincus, and Louis-Philippe Morency. 2016.
\newblock \href {https://doi.org/10.1109/MIS.2016.94} {{MOSI}: Multimodal corpus of sentiment intensity and subjectivity analysis in online opinion videos}.
\newblock \emph{IEEE Intelligent Systems}, 31(6):82--88.

\bibitem[{Zadeh et~al.(2018)Zadeh, Liang, Poria, Cambria, and Morency}]{zadeh2018cmu-mosei}
AmirAli~Bagher Zadeh, Paul~Pu Liang, Soujanya Poria, Erik Cambria, and Louis-Philippe Morency. 2018.
\newblock \href {https://doi.org/10.18653/v1/P18-1208} {Multimodal language analysis in the wild: {CMU-MOSEI} dataset and interpretable dynamic fusion graph}.
\newblock In \emph{Proceedings of the 56th Annual Meeting of the Association for Computational Linguistics (Volume 1: Long Papers)}, pages 2236--2246. Association for Computational Linguistics.

\bibitem[{Zhang et~al.(2025)Zhang, Cheng, Deng, Li, Lian, Chen, Liu, Wang, Zhang, Zhang, Guo, Zhu, Wu, Wang, Zheng, Peng, Wu, Wang, Li, Ye, and Heng}]{zhang2025mmeemotion}
Fan Zhang, Zebang Cheng, Chong Deng, Haoxuan Li, Zheng Lian, Qian Chen, Huadai Liu, Wen Wang, Yi-Fan Zhang, Renrui Zhang, Ziyu Guo, Zhihong Zhu, Hao Wu, Haixin Wang, Yefeng Zheng, Xiaojiang Peng, Xian Wu, Kun Wang, Xiangang Li, and 2 others. 2025.
\newblock \href {https://arxiv.org/abs/2508.09210} {{MME-Emotion}: A holistic evaluation benchmark for emotional intelligence in multimodal large language models}.
\newblock \emph{arXiv preprint arXiv:2508.09210}.

\bibitem[{Zhang et~al.(2023)Zhang, Li, and Bing}]{zhang2023videollama}
Hang Zhang, Xin Li, and Lidong Bing. 2023.
\newblock \href {https://aclanthology.org/2023.emnlp-demo.49/} {{Video-LLaMA}: An instruction-tuned audio-visual language model for video understanding}.
\newblock In \emph{Proceedings of the 2023 Conference on Empirical Methods in Natural Language Processing: System Demonstrations}, pages 543--553. Association for Computational Linguistics.

\end{thebibliography}


\cleardoublepage
\appendix

\begin{table*}[t]
  \centering
  \small
  \setlength{\tabcolsep}{4pt}
  \renewcommand{\arraystretch}{1.15}
  \begin{tabular}{l|ccccc|ccccc}
    \hline
      & \multicolumn{5}{c|}{\textbf{CMU-MOSI}} & \multicolumn{5}{c}{\textbf{CH-SIMS v2.0}} \\
    \hline
    \textbf{Modality} & \textbf{N} & \textbf{WN} & \textbf{Neu} & \textbf{WP} & \textbf{P}
                      & \textbf{N} & \textbf{WN} & \textbf{Neu} & \textbf{WP} & \textbf{P} \\
    \hline
    GT & 408 & 491 & 232 & 462 & 407 & 431 & 433 & 304 & 433 & 399 \\
    T  & 290 & 419 & 298 & 510 & 483 & 321 & 442 & 472 & 462 & 303 \\
    A  &  78 & 498 & 483 & 775 & 166 & 373 & 449 & 378 & 556 & 244 \\
    V  &  77 & 385 & 738 & 503 & 297 & 372 & 411 & 404 & 399 & 414 \\
    \hline
  \end{tabular}
  \caption{Five-class emotion distributions on CMU-MOSI and CH-SIMS v2.0 (each $N=2000$).}
  \label{tab:five_class_dist}
\end{table*}

\begin{figure*}[t] 
  \centering
  
  \includegraphics[width=\textwidth]{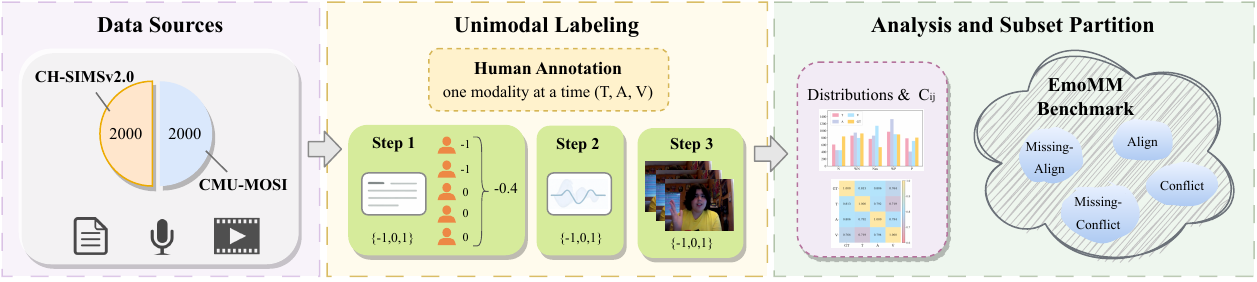}
  
  \caption{Overview of the EmoMM construction pipeline: data curation for CH-SIMS v2.0 and CMU-MOSI, single-modality human labeling for independent unimodal annotations on CMU-MOSI (CH-SIMS v2.0 provides unimodal labels), followed by statistical analysis and subset partitioning under sentiment conflicts and modality missingness.}
  \label{fig:flow}
  
\end{figure*}

\label{app:cij_matrices}

\begin{table*}[t]
  \centering
  \small
  \setlength{\tabcolsep}{4pt}
  \renewcommand{\arraystretch}{1.15}

  \begin{tabular}{l|cccc}
    \hline
    \multicolumn{5}{c}{\textbf{CMU-MOSI}} \\
    \hline
      & \textbf{GT} & \textbf{T} & \textbf{A} & \textbf{V} \\
    \hline
    \textbf{GT} & 1.0000 & 0.7998 & 0.7265 & 0.6541 \\
    \textbf{T}  & 0.7998 & 1.0000 & 0.7620 & 0.6771 \\
    \textbf{A}  & 0.7265 & 0.7620 & 1.0000 & 0.7386 \\
    \textbf{V}  & 0.6541 & 0.6771 & 0.7386 & 1.0000 \\
    \hline
  \end{tabular}
  \hspace{6pt}
  \begin{tabular}{l|cccc}
    \hline
    \multicolumn{5}{c}{\textbf{CH-SIMS v2.0}} \\
    \hline
      & \textbf{GT} & \textbf{T} & \textbf{A} & \textbf{V} \\
    \hline
    \textbf{GT} & 1.0000 & 0.8260 & 0.8854 & 0.8735 \\
    \textbf{T}  & 0.8260 & 1.0000 & 0.8219 & 0.7602 \\
    \textbf{A}  & 0.8854 & 0.8219 & 1.0000 & 0.8494 \\
    \textbf{V}  & 0.8735 & 0.7602 & 0.8494 & 1.0000 \\
    \hline
  \end{tabular}

  \caption{Cross-modal agreement coefficients $C_{ij}$.}
  \label{tab:cij}
\end{table*}

\section{Elaboration on Benchmark Construction and Analysis}
\subsection{Data Curation and Labeling}

The \textbf{EmoMM} benchmark is constructed by integrating and standardizing two widely acknowledged datasets: the English CMU-MOSI and the Chinese CH-SIMS v2.0. A primary challenge in this unification is the discrepancy in original annotation schemes and score distributions.

CH-SIMS v2.0 natively provides annotations in the range of $[-1.0, +1.0]$. We adopt its original five-class definition as the reference standard for our benchmark:
Negative ($\{-1.0, -0.8\}$), Weakly Negative ($\{-0.6, -0.4, -0.2\}$), Neutral ($\{0.0\}$), Weakly Positive ($\{0.2, 0.4, 0.6\}$), and Positive ($\{0.8, 1.0\}$).

CMU-MOSI originally provides continuous sentiment intensity scores in the range of $[-3, +3]$. To unify the formats, we first linearly scaled these scores to $[-1, +1]$ and discretized them to the nearest grid points with a step of $0.2$. 
However, simply applying the exact CH-SIMS thresholds to CMU-MOSI resulted in a significant class imbalance due to differences in the underlying data distributions. To ensure that the five-class label distribution of the English subset remains quantitatively similar to that of the Chinese subset, we adjusted the mapping boundaries slightly for CMU-MOSI. Specifically, the values $\pm 0.6$ were shifted to the extreme classes. The final aligned mapping for CMU-MOSI is:
Negative ($\{-1.0, -0.8, -0.6\}$), Weakly Negative ($\{-0.4, -0.2\}$), Neutral ($\{0.0\}$), Weakly Positive ($\{0.2, 0.4\}$), and Positive ($\{0.6, 0.8, 1.0\}$).
This rigorous alignment process guarantees that both subsets share comparable granular sentiment definitions while maintaining balanced sample sizes across categories.
Regarding the human annotation quality for the unimodal labels on CMU-MOSI, the five annotators achieved substantial consistency, with a Fleiss' Kappa score of approximately $0.54$.

To foster a comprehensive evaluation of multimodal robustness, particularly under conflicting scenarios, we employed a stratified sampling strategy rather than random selection. For the CH-SIMS subset, we select 1,000 Conflict samples (where at least one modality opposes the ground truth polarity), 500 Equal samples (full agreement across modalities), and 500 Same-Polarity-Not-Equal samples (consistent polarity but differing intensity). For the CMU-MOSI subset, we selected 2,000 samples to maintain a balanced distribution across the five classes after the aforementioned label mapping. The detailed label statistics for the textual, acoustic, and visual modalities, along with the multimodal ground truth, are summarized in Table~\ref{tab:five_class_dist}.

\subsection{Human Annotation Protocol and Ethical Considerations}
\label{app:human_annotation}

\begin{tcolorbox}[
  title=Annotator Guideline,
  colback=gray!3,
  colframe=black!55,
  boxrule=0.6pt,
  arc=2pt,
  left=6pt,right=6pt,top=6pt,bottom=6pt
]
Annotators will label sentiment for 2{,}000 aligned samples, where each instance contains three modalities (T/A/V) matched by the same \texttt{seq} identifier (video in \texttt{Video/}, audio in \texttt{Audio/}, and transcript in the spreadsheet \texttt{text} column; always verify \texttt{seq} matches the clip \texttt{name}). Use exactly one label from $\{-1,0,1\}$ for each target modality, where $-1$ indicates negative, $0$ neutral, and $1$ positive. To avoid cross-modal anchoring, complete one modality column per pass (label all T first, then all A, then all V) and do not label T/A/V consecutively for the same \texttt{seq}. For Text (T), judge from lexical content and translate if needed for accurate understanding; for Audio (A), focus on prosody such as intonation and speaking style and avoid using the transcript as the primary cue; for Video (V), mute the audio and judge sentiment only from facial expression, gesture, and visible actions. All materials are confidential: do not redistribute media/text, do not attempt identification, and do not copy content outside the secured workspace.
\end{tcolorbox}

\noindent For CMU-MOSI, we obtained independent unimodal sentiment labels through manual annotation (CH-SIMS v2.0 already provides unimodal labels). Each instance contains aligned Text (T), Audio (A), and Video (V) linked by a unique \texttt{seq} identifier (video in \texttt{Video/}, audio in \texttt{Audio/}, and transcript in the spreadsheet), and annotators were instructed to complete labeling one modality per pass (finish an entire column before switching) to reduce cross-modal anchoring and avoid information leakage across modalities. Annotators were recruited from a university student pool via lab announcements, completed standardized training plus a short calibration round, worked in teams of five with \texttt{seq}-range assignments for traceability, and were compensated at a reasonable hourly rate (about USD \$12/hour equivalent) based on time/completion rather than label outcomes. All annotators provided consent and followed confidentiality requirements: access was restricted to project storage, redistribution and any attempt to identify individuals were prohibited, and we store/release only anonymized identifiers (\texttt{seq} and clip name) with labels while minimizing potentially identifying information. Since the work involves minimal-risk labeling of existing benchmark data without new data collection from subjects, formal IRB approval is often not required under many institutional frameworks, and we adhere to standard research-ethics and data-protection practices.

\subsection{Detailed Construction of EmoMM-Missing}
\label{app:missing_construction}
To construct the EmoMM-Missing subset, we employ a controlled single-modality masking strategy.
To avoid split leakage, we first split the original full-modality clips by grouping samples with the same original sample id, and then perform the one-to-three expansion within each split.
Specifically, for each full-modality sample consisting of Text (T), Audio (A), and Video (V), we generate three distinct incomplete instances by masking exactly one modality at a time (yielding input combinations: $\{T, A\}$, $\{T, V\}$, and $\{A, V\}$).
This one-to-three expansion ensures that the model's robustness is systematically evaluated against all possible single-modality missingness scenarios for every data point.


\begin{figure}[!t]
  \centering
  \begin{subfigure}[t]{0.48\linewidth}
    \centering
    \includegraphics[height=.68\linewidth]{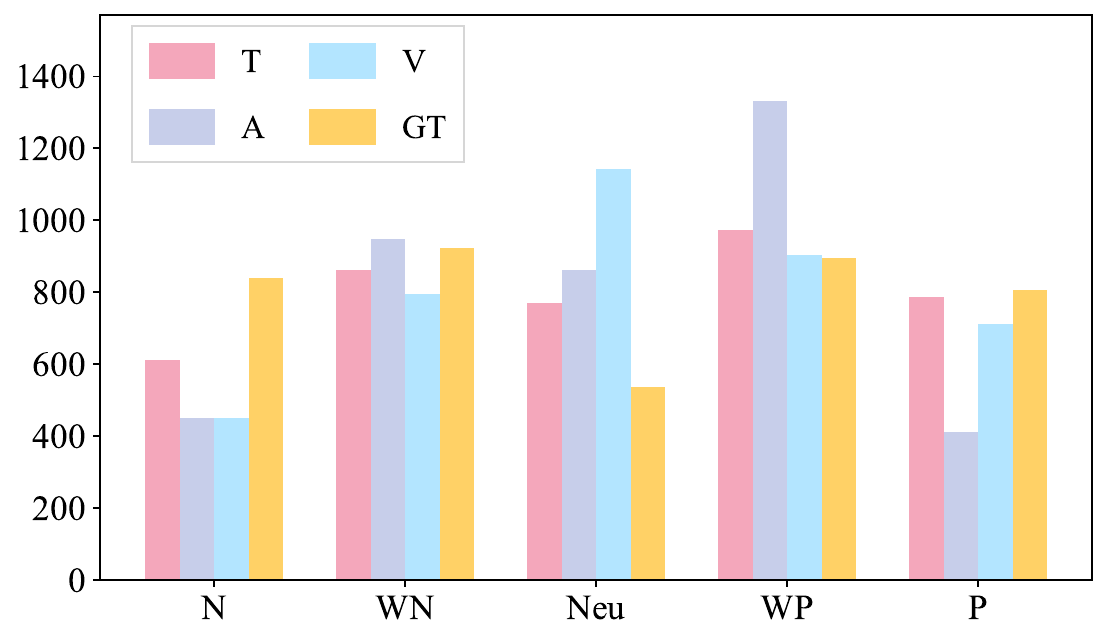}
  \end{subfigure}\hfill
  \begin{subfigure}[t]{0.48\linewidth}
    \centering
    \includegraphics[height=.68\linewidth]{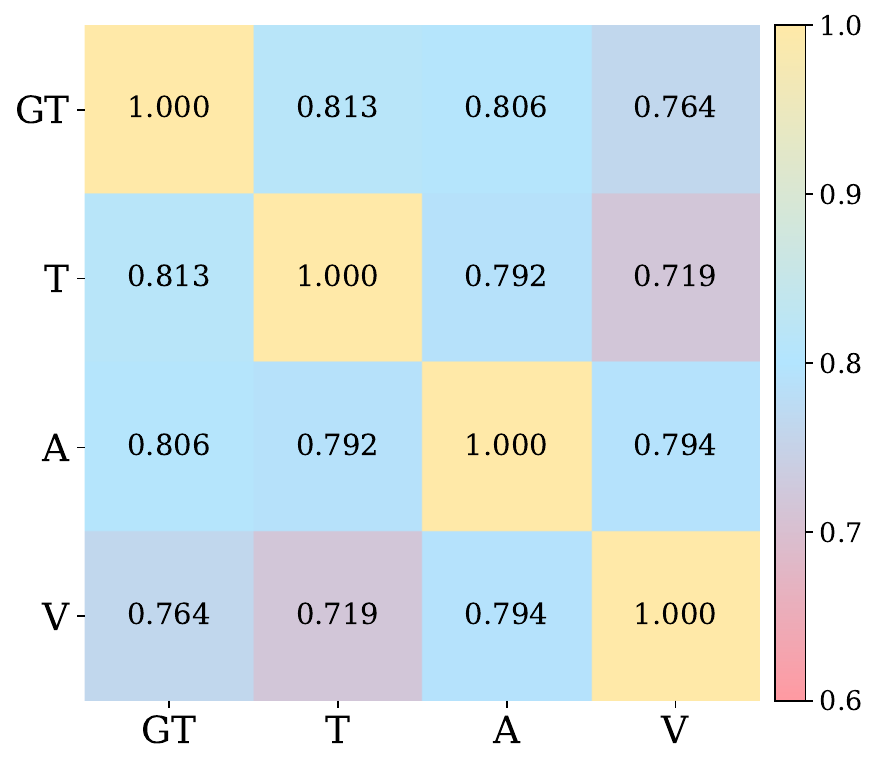}
  \end{subfigure}
  \caption{EmoMM statistics: five-class label distributions (left) and cross-modal agreement coefficients $C_{ij}$ (right), where larger $C_{ij}$ values indicate stronger agreement.}
  \label{fig:EmoMM_statistics}
\end{figure}

\begin{figure}[t]
  \centering
  \begin{subfigure}[t]{0.48\linewidth}
    \centering
    \includegraphics[height=.68\linewidth]{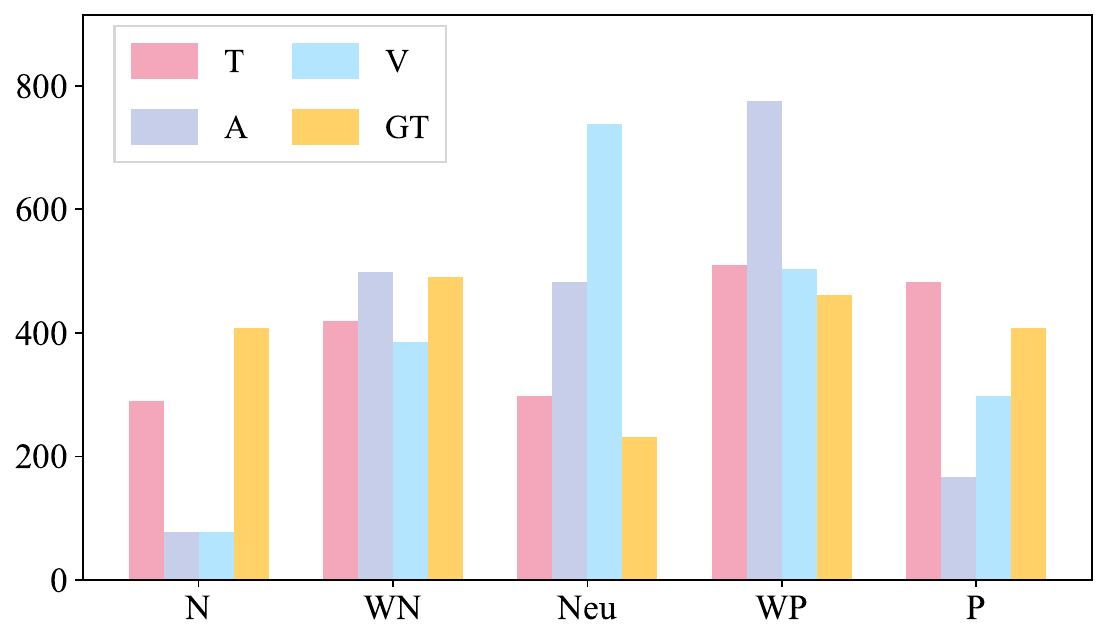}
  \end{subfigure}\hfill
  \begin{subfigure}[t]{0.48\linewidth}
    \centering
    \includegraphics[height=.68\linewidth]{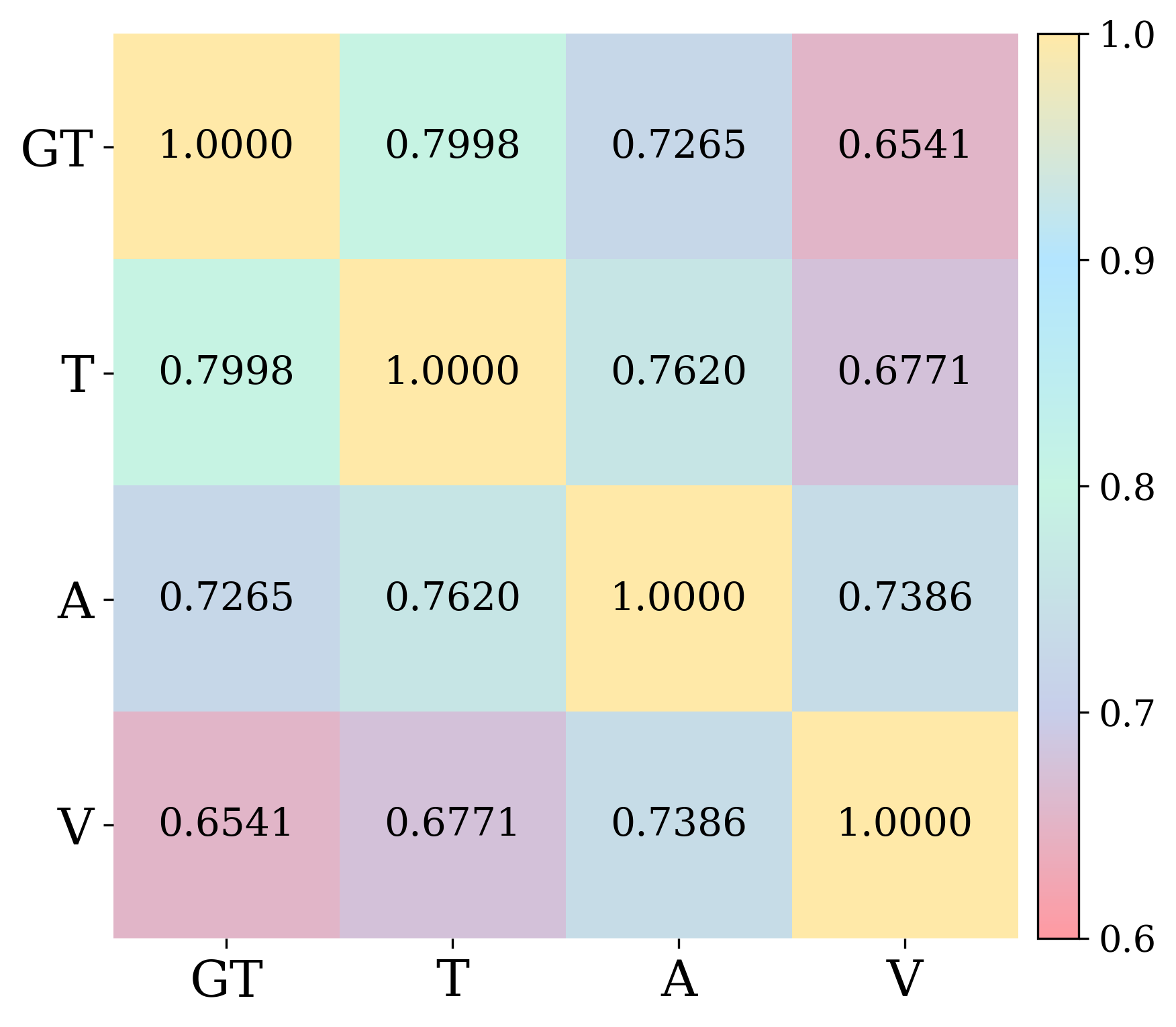}
  \end{subfigure}
  \caption{CMU-MOSI distributions and cross-modal agreement coefficients $C_{ij}$. The left and right panels correspond to those in Figure~\ref{fig:EmoMM_statistics}.}
  \label{fig:MOSI_statistics}
\end{figure}

\label{app:chsims}

\begin{figure}[t]
  \centering
  \begin{subfigure}[t]{0.48\linewidth}
    \centering
    \includegraphics[height=.68\linewidth]{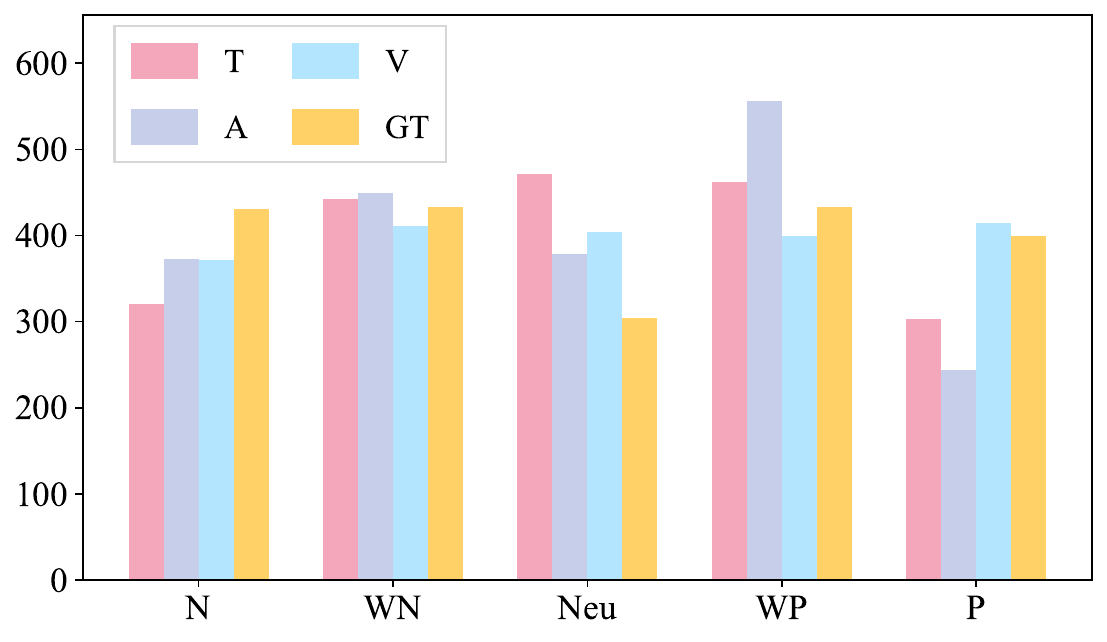}
  \end{subfigure}\hfill
  \begin{subfigure}[t]{0.48\linewidth}
    \centering
    \includegraphics[height=.68\linewidth]{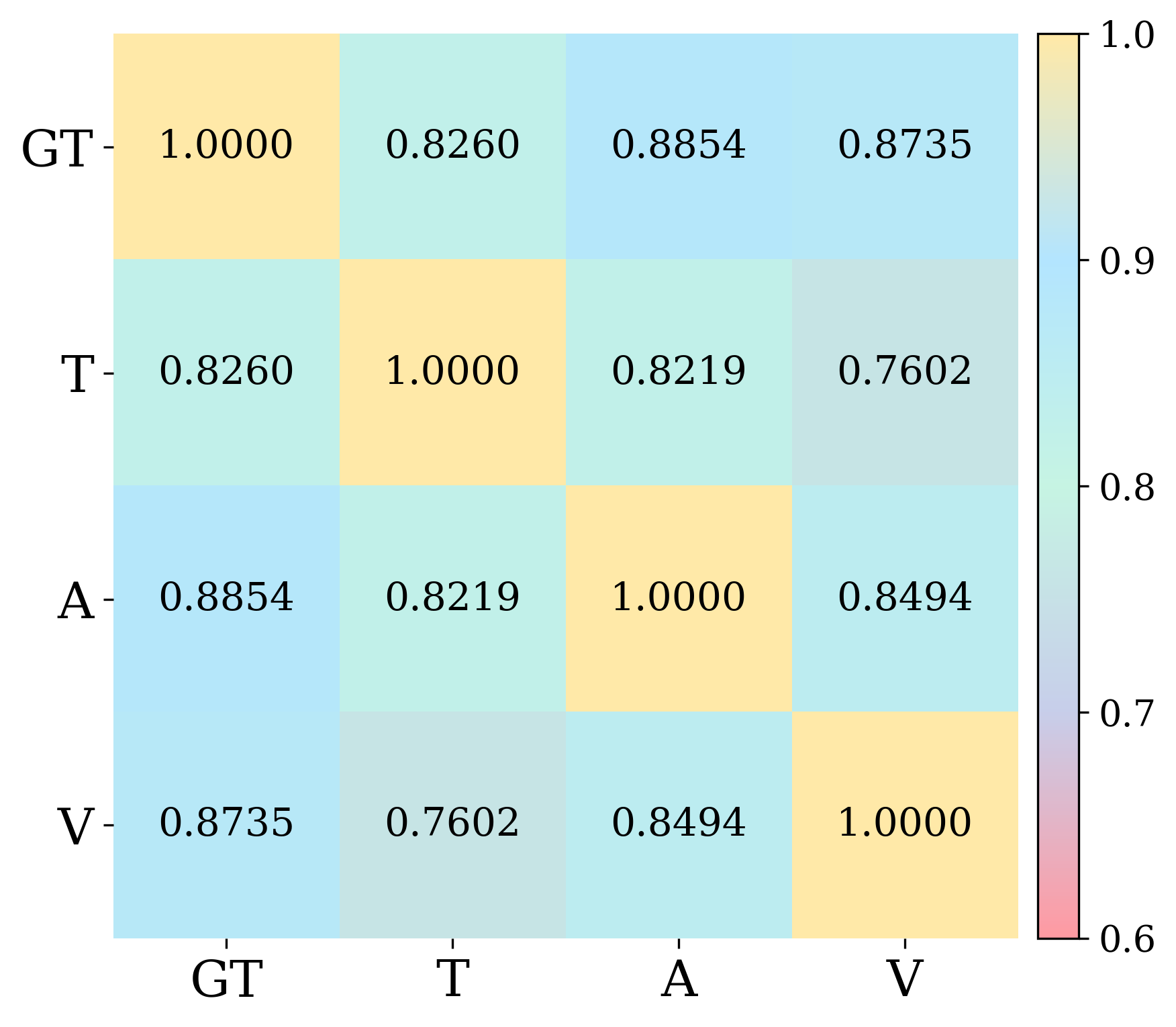}
  \end{subfigure}
  \caption{CH-SIMS v2.0 distributions and cross-modal agreement coefficients $C_{ij}$. The left and right panels correspond to those in Figure~\ref{fig:EmoMM_statistics}.}
  \label{fig:chsims_statistics}
\end{figure}

\subsection{Detailed Data Analysis}
\label{app:data_analysis}
Table~\ref{tab:five_class_dist} summarizes the count distributions of the five sentiment classes across all modalities (Text, Audio, Video) and the multimodal ground truth (GT) for both datasets.
As illustrated in Figure~\ref{fig:EmoMM_statistics} (left) and the detailed histograms in Figure~\ref{fig:MOSI_statistics} (left) and Figure~\ref{fig:chsims_statistics} (left), the distributions reveal distinct characteristics between modalities. The Audio and Video modalities tend to exhibit a higher frequency of Neutral labels compared to Text and GT, suggesting that unimodal non-verbal signals are often more subtle than lexical content.

We quantify the pairwise consistency between annotations using the metric $C_{ij}$ defined in Eq.~\ref{eq:cij} of the main text. 
Table~\ref{tab:cij} reports the full cross-modal agreement matrices for CMU-MOSI and CH-SIMS v2.0. Additionally, we visualize these relationships via heatmaps in Figure~\ref{fig:EmoMM_statistics} (right) for the combined benchmark, and individually in Figure~\ref{fig:MOSI_statistics} (right) and Figure~\ref{fig:chsims_statistics} (right). A higher $C_{ij}$ indicates stronger agreement.

As shown in Table~\ref{tab:cij} (right), all unimodal labels in the Chinese subset show high agreement with the multimodal Ground Truth (GT), demonstrating high annotation quality. Specifically, the Audio modality exhibits the smallest difference with GT, suggesting that acoustic cues play a significant role in Chinese sentiment expression in this dataset. Conversely, the largest divergence is observed between Visual and Text modalities. This aligns with the conclusions of the original paper, highlighting the frequent presence of incongruence between text and video.

For the English subset (Table~\ref{tab:cij}, left), the Text modality shows the highest correlation with GT, reinforcing the dominance of lexical cues in traditional sentiment analysis datasets. Similar to the Chinese subset, the agreement between audio and video is relatively lower, reflecting the heterogeneous nature of multimodal signals.

\subsection{Benchmark Comparison}
\label{app:benchmark_comparison}
\begin{table*}[t]
\centering
\small
\setlength{\tabcolsep}{6pt}
\renewcommand{\arraystretch}{1.12}
\begin{tabular}{lcccc}
\toprule
\textbf{Benchmark Capability} &
\textbf{MER-UniBench} &
\textbf{CA-MER} &
\textbf{MissRAG} &
\textbf{EmoMM (Ours)} \\
\midrule
Unified MLLM Evaluation Protocol      & \cmark & \cmark & \cmark & \cmark \\
Explicit Modality-Conflict Split                         & \xmark & \cmark & \xmark & \cmark \\
Explicit Modality-Missingness Split                          & \xmark & \xmark & \cmark & \cmark \\
Joint Missingness$\times$Conflict Setting  & \xmark & \xmark & \xmark & \cmark \\
Attention-level Mechanism Diagnosis         & \xmark & \cmark & \xmark & \cmark \\
Quantitative Modality Contribution / Reliance    & \xmark & \cmark & \cmark & \cmark \\
\bottomrule
\end{tabular}
\vspace{-0.5em}
\caption{Capability comparison of EmoMM against recent related benchmarks/suites for multimodal emotion understanding.}
\label{tab:benchmark_compare}
\end{table*}

Table~\ref{tab:benchmark_compare} provides a capability-level comparison between EmoMM and three recent, closely related benchmarks/suites (CA-MER, MER-UniBench, and MissRAG). The goal is to clarify what is supported by each canonical benchmark design and its released evaluation setting, rather than what could be reproduced by substantial re-engineering. 

The “Unified MLLM Evaluation Protocol” row indicates that the benchmark/suite provides a reproducible procedure to evaluate MLLMs under a consistent interface. “Explicit Modality-Conflict Split” and “Explicit Modality-Missingness Split” refer to whether conflict cases or missing-modality conditions are curated as dedicated, named stress-test settings, instead of appearing only incidentally. “Joint Missingness$\times$Conflict Setting” is stricter: missingness and conflict co-occur by design, enabling diagnosis under the hardest real-world conditions. “Attention-level Mechanism Diagnosis” is used in a narrow sense: the benchmark work reports attention/head-level analyses or attention-level interventions to explain model behavior under its stress tests. Finally, “Quantitative Modality Contribution / Reliance” refers to quantitative reporting that exposes modality reliance beyond a single aggregate score.

\section{Detailed Explanation of Metrics}

In this section, we provide a comprehensive breakdown of the three core metrics used in our analysis: Performance Shapley Modality Value (PSMV), Ground-Truth Aligned Rate (GTAR), and Normalized Mean Attention Score (nMAS). For each metric, we detail its mathematical formulation and provide a concrete example to illustrate its calculation and physical meaning.

\subsection{Performance Shapley Modality Value (PSMV)}
The Performance Shapley Modality Value (PSMV) is derived from cooperative game theory. It quantifies the average marginal contribution of a specific modality to the model's overall performance across all possible modality coalitions. By treating modalities as players in a cooperative game and the model's accuracy as the payout, PSMV provides a fair distribution of the "credit" for the prediction performance.

Formally, let $\mathcal{M}=\{T,A,V\}$ be the set of all modalities. For a specific modality $m \in \mathcal{M}$, its PSMV is calculated by averaging its marginal contribution—defined as the performance gain when adding $m$ to a subset $\mathcal{S}$—weighted by the probability of that subset occurring. The weight $w_{\mathcal{S}}$ ensures that coalitions of different sizes are balanced effectively.

\noindent\textbf{Example (PSMV for Visual Modality).}
Consider calculating the contribution of the visual modality ($V$). The possible subsets (coalitions) of $\mathcal{M}$ that do not strictly contain $V$ are $\varnothing$ (no modality), $\{T\}$, $\{A\}$, and $\{T, A\}$. Adding $V$ to these subsets yields the coalitions $\{V\}$, $\{T, V\}$, $\{A, V\}$, and $\{T, A, V\}$. Accordingly, the PSMV for $V$ is computed as:
\begin{equation}
\begin{aligned}
\mathrm{PSMV}(V)
=&~\frac{1}{3}\Big(V(\{V\})-V(\varnothing)\Big) \\
&+ \frac{1}{6}\Big(V(\{T,V\})-V(\{T\})\Big) \\
&+ \frac{1}{6}\Big(V(\{A,V\})-V(\{A\})\Big) \\
&+ \frac{1}{3}\Big(V(\{T,A,V\})-V(\{T,A\})\Big),
\end{aligned}
\end{equation}
where $V(\cdot)$ represents the model's accuracy on that specific subset. The coefficients (weights) reflect the number of permutations: for instance, there are fewer ways to form full-modality or single-modality sets compared to two-modality sets in a larger combination space. This metric reveals whether the visual modality consistently adds value (positive PSMV) or acts as noise (negative PSMV) across different contexts.

\subsection{Ground-Truth Aligned Rate (GTAR)}
While accuracy measures general performance, it does not reveal the model's decision-making preference under conflicting information. The Ground-Truth Aligned Rate (GTAR) is designed to assess the model's reliability in such scenarios. It specifically measures the conditional probability that the model makes a correct prediction given that a specific modality is trustworthy (aligned with the ground truth).

We consider a sample $x$ with available modalities $\mathcal{M}_x$. We define the \textit{Target-Conflict} set, denoted as $\mathcal{S}_{tc}$, which consists of samples where the semantic cues are contradictory. Specifically, $\mathcal{S}_{tc}$ includes samples where exactly one available modality contradicts the ground truth (e.g., in a $\{T, V\}$ sample, Text is Positive, Video is Negative, and Ground Truth is Negative). To formalize this, we utilize two indicator variables: $a_x^{m}$, which equals 1 if modality $m$ is both present and aligned with the ground truth; and $c_x$, which equals 1 if the model's final prediction matches the ground truth.

\noindent\textbf{Example (GTAR for Visual Modality).}
To calculate $\mathrm{GTAR}_V$, we focus on the subset of conflict samples where the visual modality is the "correct" one. In these cases, $a_x^{V}=1$ because the video polarity matches the ground truth ($z_x^{V} = z_x^{GT}$), while at least one other modality provides conflicting incorrect information. The metric is then computed as:
\begin{equation}
\mathrm{GTAR}_V
=
\frac{
\sum_{x\in\mathcal{S}_{tc}} a_x^{V} \cdot c_x
}{
\sum_{x\in\mathcal{S}_{tc}} a_x^{V}
}.
\label{eq:gtar_v_example}
\end{equation}
The denominator counts all conflict scenarios where the video provides the correct evidence. The numerator counts how many of those times the model actually predicted correctly. A high $\mathrm{GTAR}_V$ implies that when the video is the reliable source of information amidst conflict, the model correctly identifies and utilizes it. Conversely, a low score suggests the model ignores the video even when it holds the truth.

\begin{figure*}[t]
  \centering
  \begin{minipage}[t]{0.62\textwidth}
    \centering
    \includegraphics[width=0.9\linewidth]{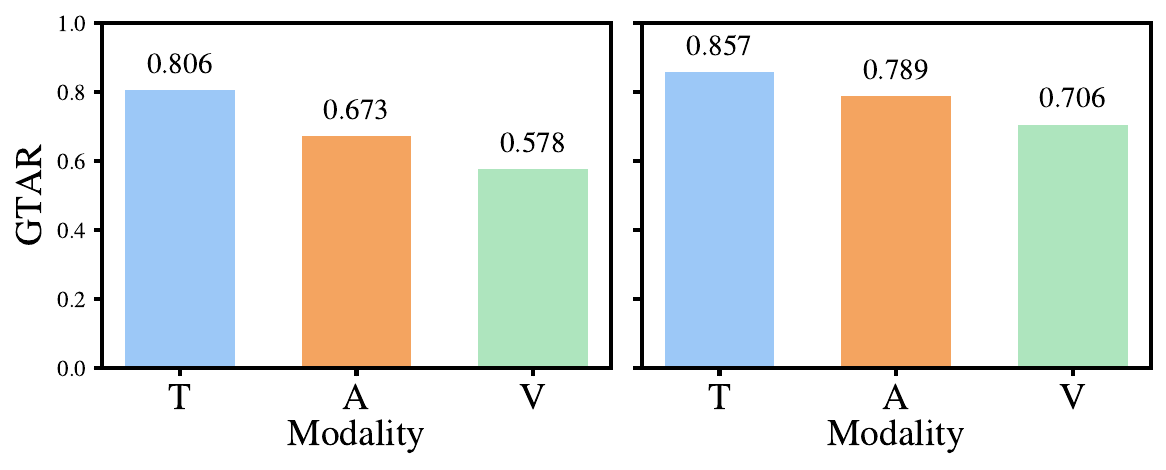}

    \vspace{-1mm}
    {\small (a)}
  \end{minipage}
  \hfill%
  \begin{minipage}[t]{0.35\textwidth}
    \centering
    \includegraphics[width=0.9\linewidth]{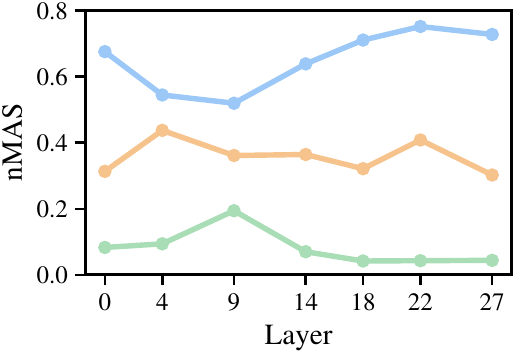}

    \vspace{-1mm}
    {\small (b)}
  \end{minipage}

  \vspace{-1mm}
  \caption{(a) GTAR on EmoMM-Missing-Conflict (left) and EmoMM-Conflict (right). (b) nMAS on EmoMM-Conflict.}
  \label{fig:gtar_nmas}
\end{figure*}

\subsection{Normalized Mean Attention Score (nMAS)}
To investigate the internal mechanism driving the model's choices, we analyze the attention distribution. However, direct summation of attention weights is biased by modality length; video inputs typically contain significantly more tokens than text or audio, naturally diluting the attention weight per token. The Normalized Mean Attention Score (nMAS) addresses this by calculating the average attention density received by the tokens of a specific modality.

For a specific decision-making step (e.g., the generation of the first sentiment token), let $q$ be the query token. The nMAS for modality $m$ in a specific attention head $(l, h)$ is the mean of the raw attention weights assigned to all tokens $K_m$ belonging to that modality.

\section{Evidence on Video Contribution Collapse}

\subsection{Analysis of GTAR and nMAS}
\label{app:GTAR}

In this section, we first examine the GTAR presented in Figure \ref{fig:gtar_nmas}(a). The quantitative results across both \textit{EmoMM-Missing-Conflict} (left) and \textit{EmoMM-Conflict} (right) reveal a rigid hierarchy of modality preference: Text ($T$) $>$ Audio ($A$) $>$ Vision ($V$). Specifically, in the missing-conflict setting where the target modality is the sole provider of correct emotional cues, the model adopts text evidence with a high probability of $0.806$, whereas the adoption rate for vision drops significantly to $0.578$. This disparity is even more pronounced in the full conflict setting, where text dominance rises to $0.857$ compared to vision's $0.706$. These figures confirm that the model exhibits a systemic "distrust" towards the visual modality, consistently treating it as a secondary information source even when it carries the ground truth.

\begin{figure*}[t]
  \centering
  \includegraphics[width=0.9\textwidth]{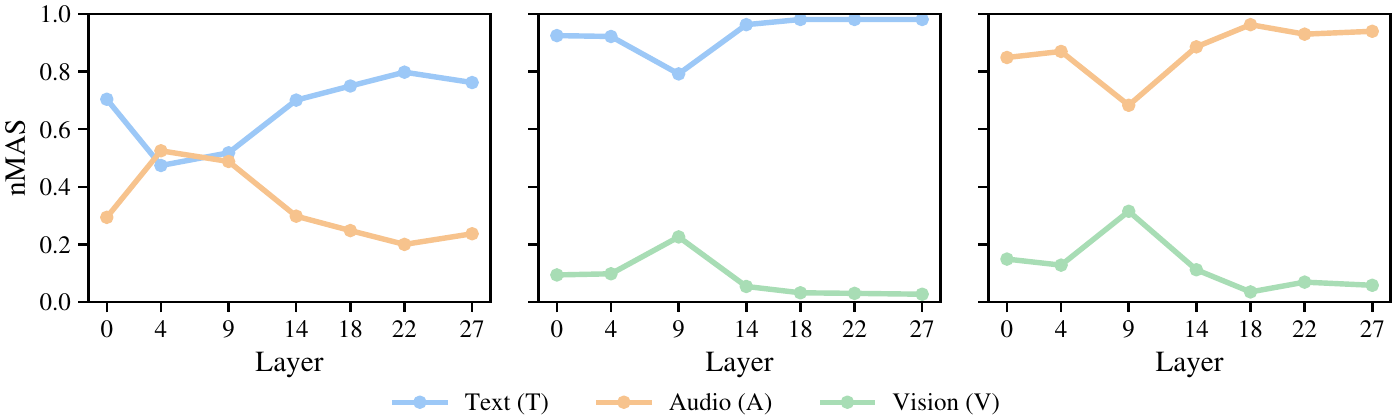}
  \caption{nMAS on EmoMM-Missing-Conflict.}

  \label{fig:nmas_missing}
\end{figure*}

To uncover the attention-level mechanism behind this preference, we further analyze the nMAS. Figure \ref{fig:nmas_missing} illustrates the pairwise attention competition in the missing-conflict scenario. In the interactions involving text and audio (left), we observe a competitive dynamic where text gradually establishes dominance in the deep layers (layers 14-27), reflecting the model's reliance on textual representations for high-level semantic reasoning. In stark contrast, whenever the visual modality is involved (center and right plots), it remains consistently suppressed. The nMAS for vision (green line) hovers at a negligible level ($<0.2$) across almost all layers, failing to gain significant attention weight against either text or audio.

\label{app:metrics_explanation}

\begin{figure}[t]
\centering
\includegraphics[width=0.9\linewidth]{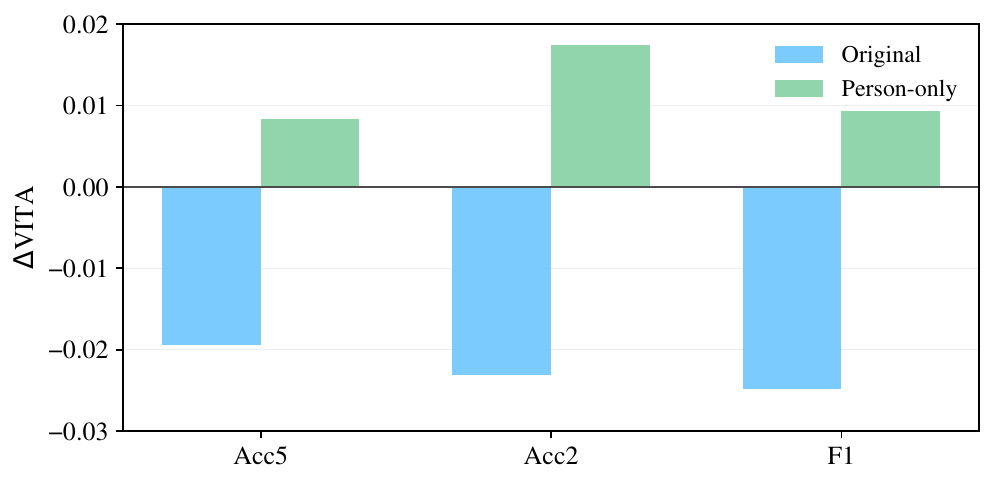}
\vspace{-2mm}
\caption{Marginal gain of video on the conflict subset. }
\label{fig:person_mask_ablation}
\vspace{-2mm}
\end{figure}

This trend is corroborated by the three-way competition dynamics shown in Figure \ref{fig:gtar_nmas}(b) for the \textit{EmoMM-Conflict} setting. Here, the text modality (blue line) ascends rapidly in the middle-to-deep layers, securing the majority of the attention budget, while the audio modality (orange line) maintains a moderate presence. The visual modality, however, stays at the bottom of the attention distribution throughout the network depth.

\begin{figure*}[t]
  \centering
  \includegraphics[width=\textwidth]{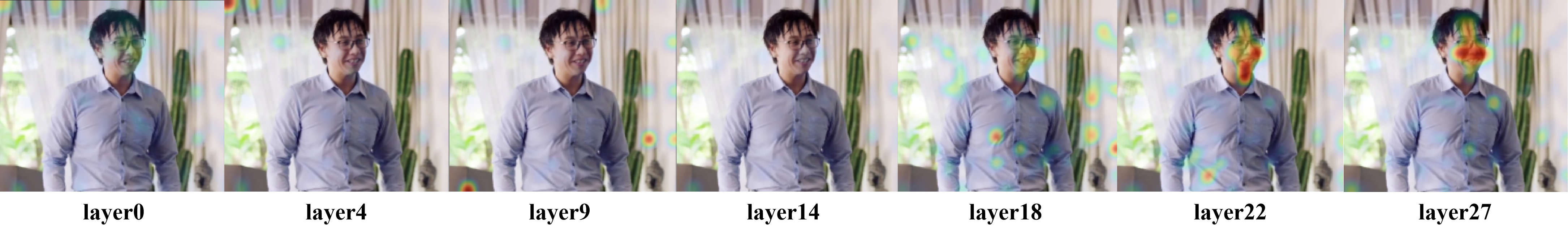}
  \caption{Heatmap of attention on the video frame (The facial images displayed in this figure are sourced from the CH-SIMS v2.0 dataset).}

  \label{fig:heatmap}
\end{figure*}

\begin{figure*}[t]
  \centering
  \includegraphics[width=\textwidth]{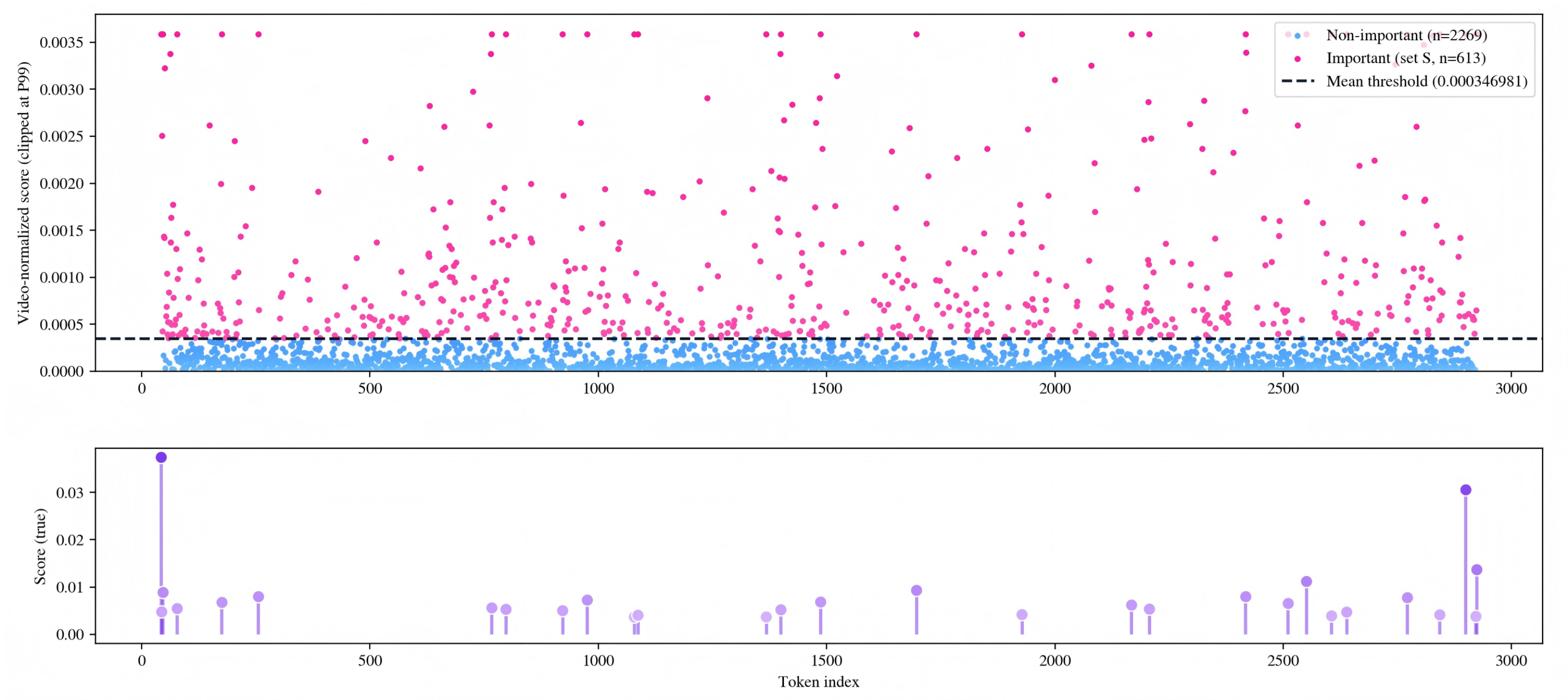}
  \caption{Importance Score of video tokens.}

  \label{fig:video_token_mean}
\end{figure*}


\subsection{Additional Validation: Person-region Masking Improves Video Evidence Density}
\label{app:person_mask}

In Sec.\ref{sec:Pre}, we hypothesize that VCC on the conflict subset is partially caused by the large scale and redundancy of raw video tokens: background and irrelevant patches dilute informative evidence, leading to suboptimal attention allocation and a non-positive marginal gain when adding video to \texttt{TA}.

To validate this hypothesis, we localize the person region using MediaPipe and suppress background tokens outside the person/face ROI. The intervention only changes the video input/embeddings while keeping prompts, decoding, and evaluation pipeline unchanged.

We report the marginal gain of video on the conflict subset:
\begin{equation}
\Delta V = \mathrm{Perf}(\texttt{TAV}) - \mathrm{Perf}(\texttt{TA}),
\label{eq:delta_v_app}
\end{equation}
where $\mathrm{Perf}(\cdot)$ is ACC-5, ACC-2, or F1.
Figure~\ref{fig:person_mask_ablation} shows that $\Delta V$ flips from negative (Original) to positive (Person-only),
indicating that increasing video evidence density leads to significantly improved video attention utilization and restores video utility under modality conflicts.

\section{Additional Implementation Details of CHASE}

\subsection{Formula Details}
\label{app:loss_jsd}

\paragraph{Kullback--Leibler divergence.}
For two discrete distributions $P=(P_m)_{m\in\mathcal{M}}$ and $Q=(Q_m)_{m\in\mathcal{M}}$
over the modality set $\mathcal{M}$ (where $P_m\ge0$, $Q_m\ge0$, and $\sum_{m\in\mathcal{M}}P_m=\sum_{m\in\mathcal{M}}Q_m=1$),
we define
\begin{equation}
D_{\mathrm{KL}}(P\|Q)
=\sum_{m\in\mathcal{M}} P_m \log \frac{P_m}{Q_m+\epsilon},
\label{eq:kl_def}
\end{equation}
where $\log(\cdot)$ is the natural logarithm, $\epsilon>0$ is a small constant for numerical stability,
and we follow the convention that $0\log 0 = 0$.
In our Router loss (Eq.~\eqref{eq:router_loss}), we use $D_{\mathrm{KL}}(\mathbf{g}^{\star}\|\mathbf{g}^{\ell})$.

\paragraph{Binary cross-entropy (BCE).}
Given a binary label $y\in\{0,1\}$ and a predicted probability $p\in(0,1)$, we define
\begin{equation}
\mathrm{BCE}(y,p)
=-\Big(y\log(p+\epsilon) + (1-y)\log(1-p+\epsilon)\Big),
\label{eq:bce_def}
\end{equation}
where $\epsilon>0$ avoids numerical issues when $p$ is close to $0$ or $1$.
In Eq.~\eqref{eq:router_loss}, $y=p^{\star}$ and $p=p^{\ell}$.

\paragraph{Jensen--Shannon divergence (JSD).}
For two distributions $P,Q$ over $\mathcal{M}$, let $M=\frac{1}{2}(P+Q)$, i.e., $M_m=\frac{1}{2}(P_m+Q_m)$.
We define
\begin{equation}
\mathrm{JSD}(P\|Q)
=\frac{1}{2}D_{\mathrm{KL}}(P\|M)+\frac{1}{2}D_{\mathrm{KL}}(Q\|M).
\label{eq:jsd_def}
\end{equation}
In head selection (Eq.~\eqref{eq:head_select}), we set $P=\boldsymbol{\pi}^{\ell,h}$ and $Q=\mathbf{g}^{\ell}$.

\subsection{Calculation of Token Importance Score}
\label{sec:appendix_importance}

In this section, we provide the detailed formulation for the token importance score $I_k$, which is used to identify the Significant Visual Token Set ${S_v}$ in the Intra-modality Bias module.

To determine which input tokens are most critical for the sentiment classification decision, we construct an importance metric based on the model's generative process. We treat the sentiment classification task as the first step of a generative decoding process. Specifically, when the model predicts the sentiment label, it computes the probability distribution over the vocabulary. We focus on the logits corresponding to the candidate sentiment labels $\mathcal{L} = \{A, B, C, D, E\}$.

Let $z_{\text{id}(\ell)}(X)$ denote the logit value for a label character $\ell \in \mathcal{L}$ given the input $X$. The model's predicted label $\ell^*$ is determined by the maximum logit among the candidates:
\begin{equation}
    \ell^* = \operatorname*{arg\,max}_{\ell \in \{A, \dots, E\}} z_{\text{id}(\ell)}(X).
\end{equation}
We define the target scalar output $y(X)$ as the logit of this predicted label:
\begin{equation}
    y(X) = z_{\text{id}(\ell^*)}(X).
\end{equation}
To quantify the contribution of each token to this prediction, we adopt a gradient-based saliency approach. We utilize the \textit{Gradient $\times$ Input} method, which combines the sensitivity of the output with respect to the input and the activation strength. This is a standard and computationally efficient baseline in interpretability research.

Let $\mathbf{h}_k$ represent the hidden state (or embedding vector) of the $k$-th token. The importance score $I_k$ is calculated as the magnitude of the dot product between the gradient of the target output $y(X)$ with respect to $\mathbf{h}_k$ and the hidden state $\mathbf{h}_k$ itself:
\begin{equation}
    I_k = \left| \left\langle \frac{\partial y(X)}{\partial \mathbf{h}_k}, \mathbf{h}_k \right\rangle \right|.
\end{equation}
This score reflects the magnitude of the $k$-th token's influence on the final logit $y(X)$. To ensure the scores are comparable and valid for probability-like masking, we normalize the raw importance scores across all tokens in the sequence to obtain $\hat{I}_k$:
\begin{equation}
    \hat{I}_k = \frac{I_k}{\sum_{j=1}^{K_V} I_j + \epsilon},
\end{equation}
where $\epsilon$ is a small constant for numerical stability, and $K_V$ represents the set of visual tokens. These normalized scores $\hat{I}_k$ are subsequently used to threshold and select the Significant Visual Token Set $S_v$ as described in the main text.

\subsection{CHASE Algorithm}
\label{app:chase_algorithm}

Algorithm~\ref{alg:chase_shared} presents the core procedure of CHASE. 
It consists of two stages: (i) training a shared Router to predict conflict probability and modality gate targets, and (ii) performing inference-time dynamic layer/head steering based on the Router outputs.

\SetKwInput{KwInput}{Input}
\SetKwInput{KwOutput}{Output}

\begin{algorithm*}[t]
\caption{CHASE: Conflict-aware Head-level Attention Steering}
\label{alg:chase_shared}

\textbf{Stage 1: Shared Router Training}

\KwInput{Training set $\mathcal{D}_{\text{train}}$; Frozen MLLM $\mathcal{M}$ with $L$ layers and $H$ heads; Start layer $l_r$; Hyperparameters $\lambda_1,\lambda_2$.}

Initialize Router parameters $\theta_r$ (shared across layers) for conflict head and gate head\;
\While{not converged}{
    Sample batch $(x,y)$ from $\mathcal{D}_{\text{train}}$\;
    Construct conflict label $p^\star$ and target gate $\mathbf{g}^\star$ from unimodal/multimodal labels (Eq.~\eqref{eq:target_gate})\;
    \For{$\ell \leftarrow l_r$ \KwTo $L$}{
        Extract layer features and modality summaries to form Router input $\mathbf{x}^{\ell}$\;
        Compute head-wise modality preference $\boldsymbol{\pi}^{\ell,h}$ (Eq.~\eqref{eq:pi_head}) and layer preference $\boldsymbol{\pi}^{\ell}$\;
        Predict $p^{\ell}$ and $\mathbf{g}^{\ell}$ using shared Router $\theta_r$\;
        $\mathcal{L} \leftarrow \mathcal{L}
        + \lambda_1 D_{\mathrm{KL}}(\mathbf{g}^\star \| \mathbf{g}^{\ell})
        + \lambda_2 \mathrm{BCE}(p^\star, p^{\ell})$\;
    }
    Update $\theta_r$ to minimize $\mathcal{L}$ while keeping $\mathcal{M}$ frozen\;
}
\KwOutput{Trained shared Router parameters $\theta_r$.}

\vspace{0.3cm}

\textbf{Stage 2: Inference-time Dynamic Layer/Head Steering}

\KwInput{Test input $x$; Frozen MLLM $\mathcal{M}$; Trained shared Router $\theta_r$; Threshold $\tau$; Top-$K$ heads $K$; Steering hyperparameters $\eta,\gamma,\beta$.}

\For{$\ell \leftarrow 1$ \KwTo $L$}{
    Compute attention logits $\{s_{qk}^{\ell,h}\}_{h=1}^H$ in layer $\ell$\;
    \If{$\ell \ge l_r$}{
        Form Router input $\mathbf{x}^{\ell}$ and compute $(p^{\ell},\mathbf{g}^{\ell})$ using shared Router $\theta_r$\;
        Compute $\boldsymbol{\pi}^{\ell,h}$ for all heads (Eq.~\eqref{eq:pi_head})\;
        \If{$p^{\ell} > \tau$}{
            Compute head scores $s_{\ell,h}=\mathrm{JSD}(\boldsymbol{\pi}^{\ell,h}\|\mathbf{g}^{\ell})$\;
            Select conflict-critical heads $\mathcal{H}^{\ell} \leftarrow \mathrm{TopK}_{h\in\{1,\dots,H\}}\, s_{\ell,h}$\;
            \ForEach{$h \in \mathcal{H}^{\ell}$}{
                Compute target preference $\widehat{\boldsymbol{\pi}}^{\ell,h}$ (Eq.~\eqref{eq:quota})\;
                Derive inter-modality bias $b^{\ell,h}_m$ (Eq.~\eqref{eq:inter_bias})\;
                Compute intra-modality bias $\delta_{qk}^{\ell,h}$ (Eq.~\eqref{eq:intra_bias})\;
                Update logits:
                $\tilde{s}_{qk}^{\ell,h} \leftarrow s_{qk}^{\ell,h}
                + \gamma\cdot b^{\ell,h}_{m(k)}
                + \beta\cdot \delta_{qk}^{\ell,h}$\;
            }
        }
    }
    Apply softmax on logits and proceed to next layer\;
}

\KwOutput{Model prediction $\hat{y}$.}
\end{algorithm*}

\begin{table*}[t]
\centering
\small
\begin{tabular}{lccc}
\toprule
Split & Conflict Detection Recall & Target-Modality Top-1 on Steered Layers & Joint Success Rate \\
\midrule
Valid & 0.9024 & 0.7716 & 0.7018 \\
Test  & 0.8965 & 0.7790 & 0.7055 \\
\bottomrule
\end{tabular}
\caption{Conflict detection and target-modality identification results on true conflict samples at $\tau=0.7$.}
\label{tab:router_conflict_identification}
\end{table*}

\FloatBarrier

\subsection{Conflict Detection and Target-Modality Identification}
As shown in Table~\ref{tab:router_conflict_identification}, we further report the conflict detection and target-modality identification performance of the Router in CHASE on true conflict samples.
\textit{Conflict Detection Recall} measures whether true conflict samples successfully trigger conflict-related steering.
\textit{Target-Modality Top-1 on Steered Layers} measures, among samples that trigger steering, whether the gate assigns its top-1 preference to the target modality on the layers where steering is actually applied.
\textit{Joint Success Rate} measures whether conflict detection and target-modality identification succeed simultaneously on true conflict samples.
Under the threshold used in our main experiments ($\tau = 0.7$), the Router achieves high conflict-detection recall on both the validation and test splits, reaching 0.9024 and 0.8965, respectively; among true conflict samples that trigger steering, the target-modality top-1 accuracy reaches 0.7716 on validation and 0.7790 on test.

Furthermore, the joint success rate reaches 0.7018 on validation and 0.7055 on test, indicating that in about 70\% of true conflict cases, CHASE not only detects the conflict but also assigns its primary gate preference to the target modality on the layers where steering is actually applied.
These results suggest that the gains of CHASE are not merely black-box improvements in final prediction accuracy, but are consistent with its intended mechanism: the Router can reliably trigger steering in most true conflict samples and provide effective target-modality guidance for subsequent attention reallocation.

\section{Baseline Prompting \& Output-to-Label Mapping}
\label{app:prompting}

\paragraph{System prompt.}
\begin{Prompt}
    
You are a multimodal sentiment analysis assistant.
You may receive any subset of modalities (video, audio, transcript text).
Do not assume missing modalities or fabricate details not supported by the evidence.
Follow the required output format exactly and output nothing else.

\end{Prompt}

\newpage
\paragraph{User message content order.}
\begin{Prompt}
(1) {video file}   (if VIDEO is present)
(2) {audio file}   (if AUDIO is present)
(3) Transcript (may contain ASR errors):
    <TRANSCRIPT_TEXT>   (if TEXT is present)
(4) <INSTRUCTION BELOW>
\end{Prompt}

\paragraph{Instruction.}
\begin{Prompt}
Task: Based on the provided modalities (<PRESENT_MODALITIES>), assign a sentiment label and score.
Requirements:
 Decision process:
  1) First decide the Sentiment LETTER (A–E) based on all available modalities.
  2) Then choose ONE Score from the allowed set of that LETTER that best reflects the sentiment intensity.
    A (Negative) = <SET_A>, B (Weakly Negative) = <SET_B>,
    C (Neutral) = <SET_C>, D (Weakly Positive) = <SET_D>, E (Positive) = <SET_E>
Output format (exactly):
<ONE LETTER: A or B or C or D or E>
Score: <one value from the allowed set matching the letter>
\end{Prompt}

\paragraph{Score policy for CH-SIMSv2.}
\begin{Prompt}
A: {-1.0, -0.8}
B: {-0.6, -0.4, -0.2}
C: { 0.0}
D: { 0.2,  0.4,  0.6}
E: { 0.8,  1.0}
\end{Prompt}

\paragraph{Score policy for CMU-MOSI.}
\begin{Prompt}
A: {-1.0, -0.8, -0.6}
B: {-0.4, -0.2}
C: { 0.0}
D: { 0.2,  0.4}
E: { 0.6,  0.8,  1.0}
\end{Prompt}

\textbf{}

\section{Show-case: Qualitative Case Studies}
\label{app:showcase}

This section provides qualitative case studies to illustrate model behavior under modality conflict and joint missingness–conflict. All examples are drawn from EmoMM. We present each example in a unified case-card layout (see the case-card tables/figures) to enable clean side-by-side comparison between the baseline and CHASE. The case-card visually summarizes the multimodal evidence (text, audio cue, and representative video frames), the ground-truth and unimodal label, as well as the model predictions and attention-budget snapshots across layers, followed by a brief interpretation.

\definecolor{attT}{gray}{0.80} 
\definecolor{attA}{gray}{0.55} 
\definecolor{attV}{gray}{0.25} 

\newcolumntype{P}[1]{>{\raggedright\arraybackslash}p{#1}}
\newcolumntype{C}[1]{>{\centering\arraybackslash}p{#1}}

\newcommand{\AttnBar}[3]{%
\begin{tikzpicture}[x=2.45cm,y=0.22cm,baseline=-0.65ex]
  \pgfmathsetmacro{\t}{#1}
  \pgfmathsetmacro{\a}{#2}
  \draw[black, line width=0.25pt] (0,0) rectangle (1,1);
  \fill[attT] (0,0) rectangle (\t,1);
  \fill[attA] (\t,0) rectangle ({\t+\a},1);
  \fill[attV] ({\t+\a},0) rectangle (1,1);
\end{tikzpicture}%
}

\newcommand{\AttnRow}[4]{%
\texttt{#1}\hspace{0.35em}\AttnBar{#2}{#3}{#4}\hspace{0.55em}%
{\scriptsize (T=#2,\;A=#3,\;V=#4)}%
}

\newcommand{\AttnLegend}{%
{\scriptsize
\begin{tikzpicture}[x=0.85cm,y=0.18cm,baseline=-0.65ex]
  \fill[attT] (0,0) rectangle (0.22,1);
  \node[anchor=west] at (0.25,0.5)  {T};
  \fill[attA] (0.48,0) rectangle (0.70,1);
  \node[anchor=west] at (0.73,0.5)  {A};
  \fill[attV] (0.96,0) rectangle (1.18,1);
  \node[anchor=west] at (1.21,0.5)  {V};
\end{tikzpicture}}
}

\newcommand{\LabelStack}[1]{%
  \colorbox{black!6}{%
    \parbox{0.9\linewidth}{\scriptsize\ttfamily #1}%
  }%
}

\begin{table*}[!t]
\centering
\small
\setlength{\tabcolsep}{4pt}
\renewcommand{\arraystretch}{1.10}

\begin{tabular}{P{0.12\textwidth} P{0.42\textwidth} P{0.42\textwidth}}
\toprule
\multicolumn{3}{c}{\textbf{Conflict Case Analysis}} \\
\midrule

\multicolumn{3}{@{}P{\textwidth}@{}}{%
\begin{minipage}[t]{\linewidth}
\vspace{2pt}
\begin{minipage}[t]{0.32\linewidth}
\raggedright
\textit{Text:} ``Hey, got a cigarette?''\par
\vspace{2pt}
\textit{Audio:} The audio implies a desire for cigarettes, but also a somewhat casual or relaxed tone.\par
\vspace{6pt}
\LabelStack{%
Ground-Truth: label=P(1.0)\\
    T= Neu(0.0) A= WP(0.4) V= P(1.0)
}

\end{minipage}\hfill%
\begin{minipage}[t]{0.66\linewidth}
\centering
\vspace{1pt}
\begin{tabular}{@{}c@{\hspace{5pt}}c@{\hspace{5pt}}c@{}}
\includegraphics[width=0.32\linewidth]{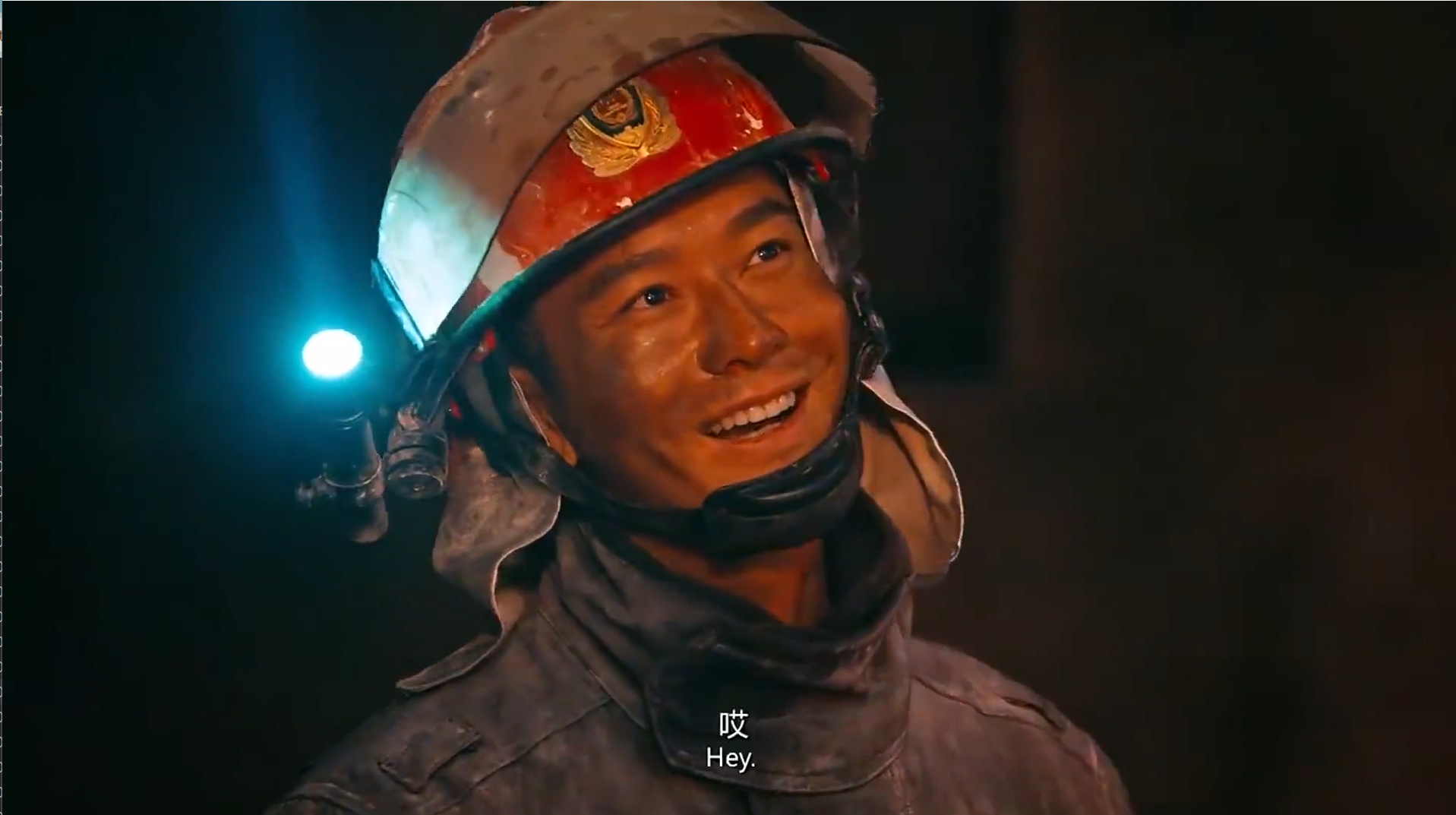} &
\includegraphics[width=0.32\linewidth]{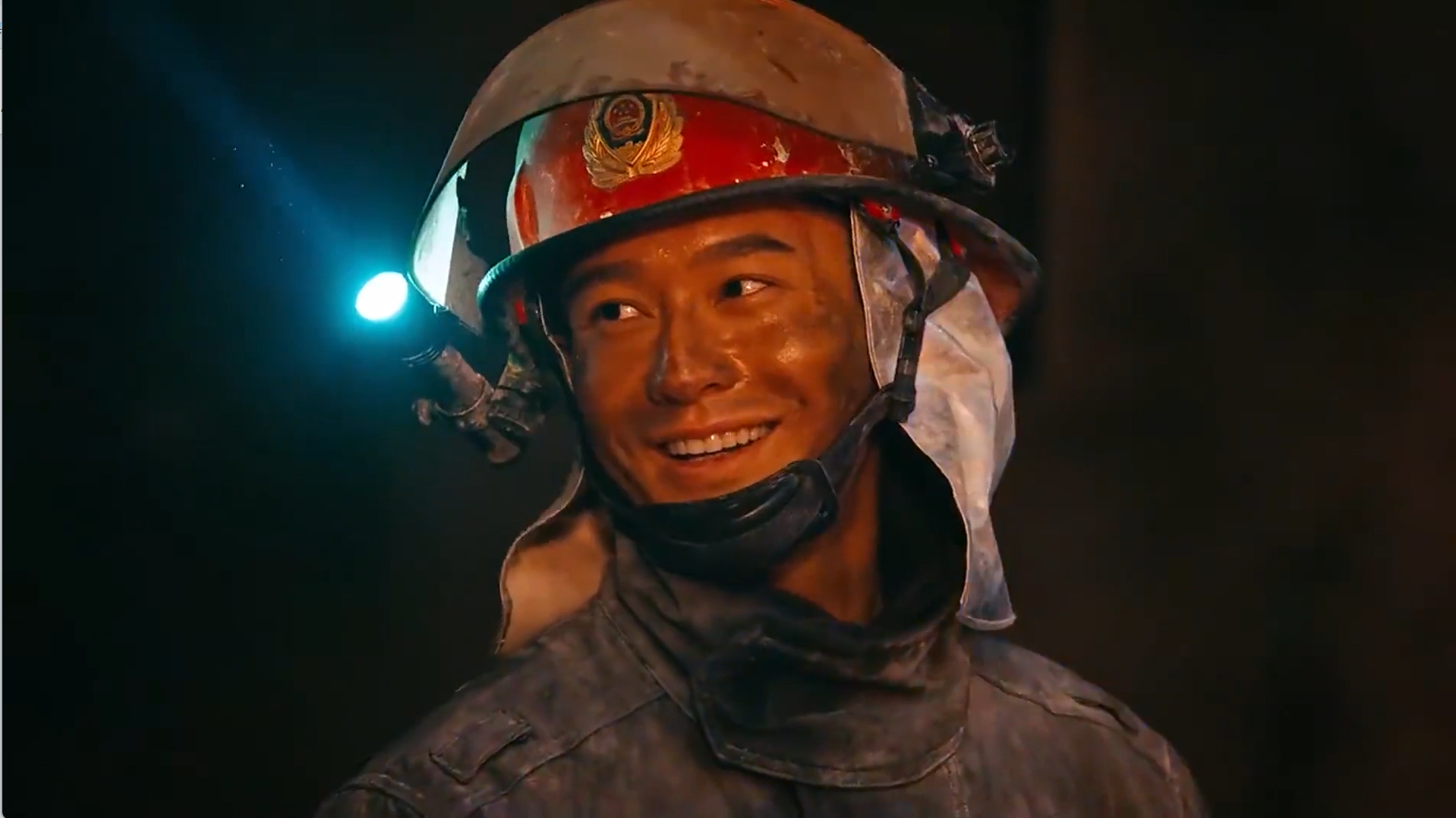} &
\includegraphics[width=0.32\linewidth]{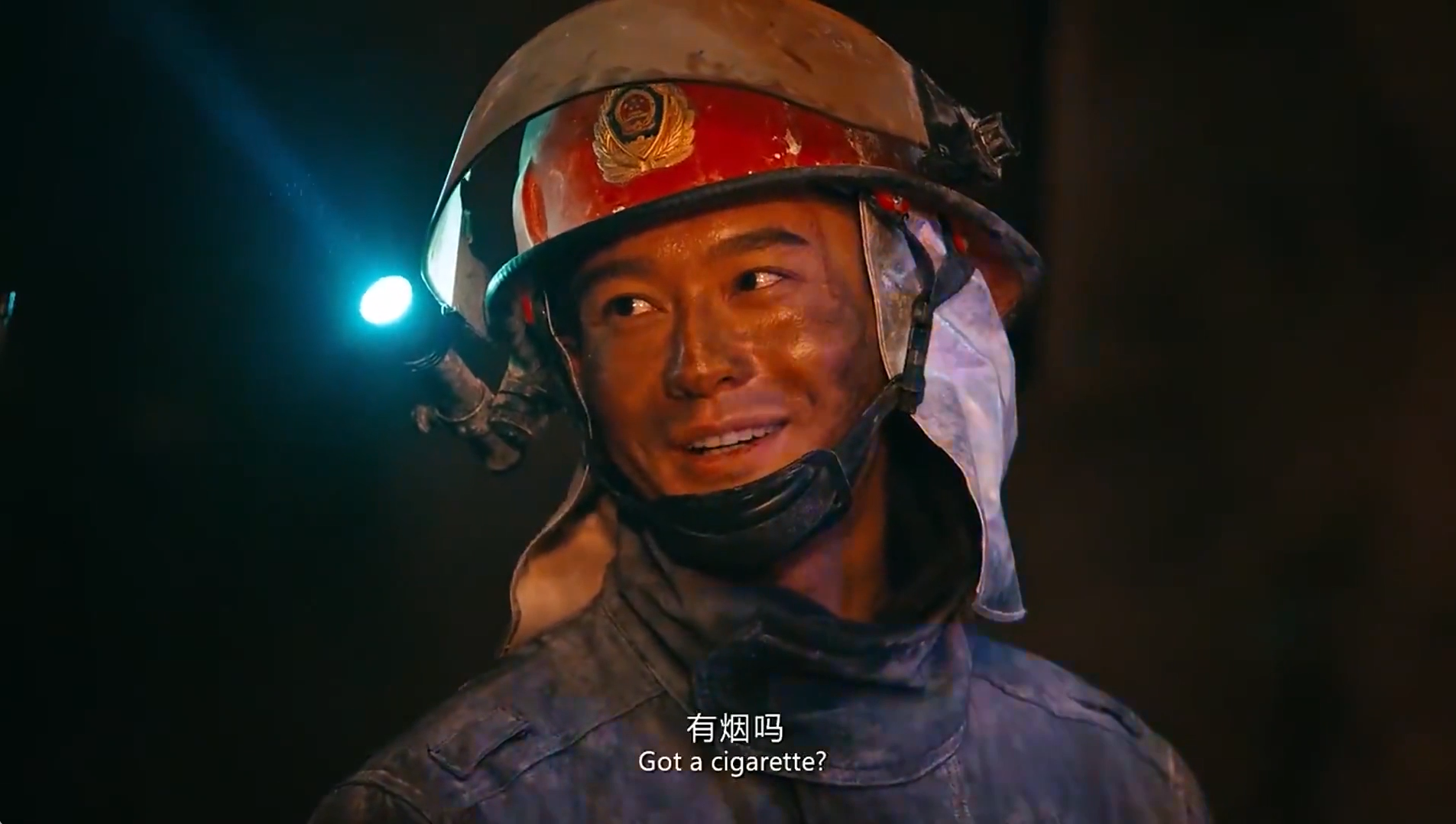} \\
{\scriptsize Frame-1} & {\scriptsize Frame-2} & {\scriptsize Frame-3}
\end{tabular}
\end{minipage}
\vspace{2pt}
\end{minipage}
} \\
\midrule

\textbf{Item} & \textbf{Baseline} & \textbf{CHASE} \\
\midrule

Prediction &
\texttt{label=N (-0.8)} \xmark &
\texttt{label=P (1.0)} \cmark \\
\midrule

Attention budget\newline\AttnLegend &
\AttnRow{L13}{0.7436}{0.2319}{0.0245}\newline
\AttnRow{L17}{0.8324}{0.1584}{0.0092}\newline
\AttnRow{L21}{0.8555}{0.1356}{0.0089}\newline
\AttnRow{L26}{0.7032}{0.2842}{0.0126}\newline
\AttnRow{L28}{0.8300}{0.1481}{0.0220}
&
\AttnRow{L13}{0.7069}{0.2470}{0.0461}\newline
\AttnRow{L17}{0.7875}{0.1802}{0.0323}\newline
\AttnRow{L21}{0.8039}{0.1633}{0.0328}\newline
\AttnRow{L26}{0.6582}{0.3047}{0.0370}\newline
\AttnRow{L28}{0.7758}{0.1751}{0.0495} \\
\midrule

Interpretation &
\multicolumn{2}{P{0.84\textwidth}}{%
\parbox[t]{0.84\textwidth}{%
\raggedright
\setlength{\emergencystretch}{2em}
\footnotesize
The unimodal labels indicate a conflict pattern (V matches GT, while T and A conflicts).
The baseline increasingly concentrates attention on audio at late layers, coinciding with the incorrect prediction.
CHASE re-balances the late-layer attention budget, recovering the GT-aligned positive prediction.
}} \\
\bottomrule
\end{tabular}

\vspace{-0.6em}
\caption{Conflict case analysis.}
\label{tab:conflict_case_final}
\end{table*}

\begin{table*}[!t]
\centering
\small
\setlength{\tabcolsep}{4pt}
\renewcommand{\arraystretch}{1.10}

\begin{tabular}{P{0.12\textwidth} P{0.42\textwidth} P{0.42\textwidth}}
\toprule
\multicolumn{3}{c}{\textbf{Missing Case Analysis}} \\
\midrule

\multicolumn{3}{@{}P{\textwidth}@{}}{%
\begin{minipage}[t]{\linewidth}
\vspace{2pt}
\begin{minipage}[t]{0.34\linewidth}
\raggedright
\textit{Text:} ``I won't give up.''\par
\vspace{2pt}
\textit{Audio:} \textbf{[missing]}\par
\vspace{6pt}

\LabelStack{%
\texttt{Ground-Truth: label= N(-1.0)}\\
\texttt{T= P(0.8) V= N(-1.0)}
}
\end{minipage}\hfill%
\begin{minipage}[t]{0.64\linewidth}
\centering
\vspace{1pt}
\begin{tabular}{@{}c@{\hspace{5pt}}c@{\hspace{5pt}}c@{}}
\includegraphics[width=0.32\linewidth]{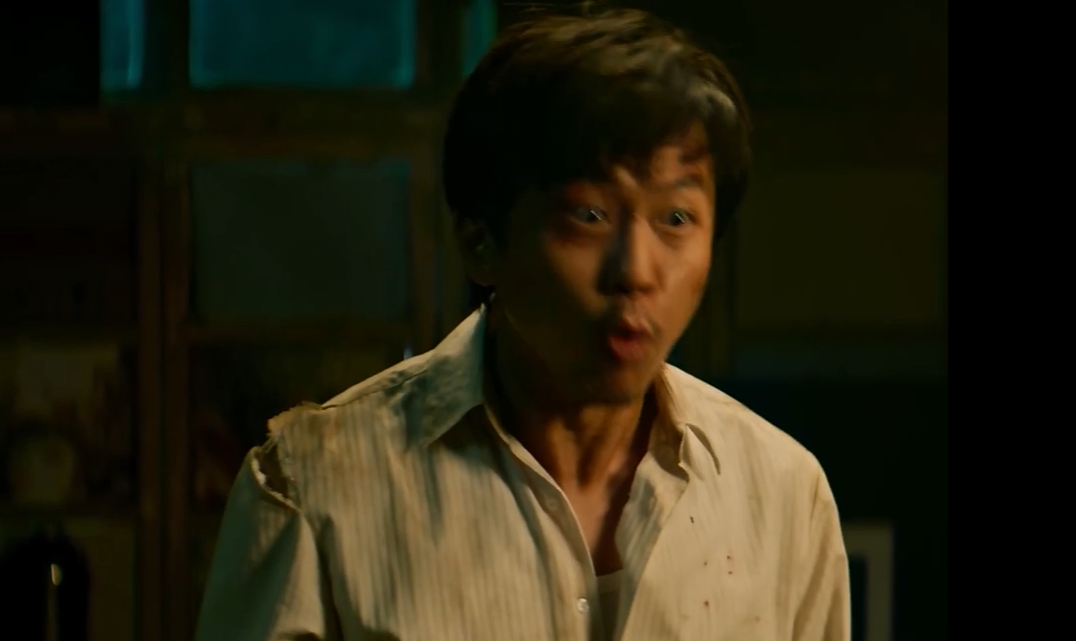} &
\includegraphics[width=0.32\linewidth]{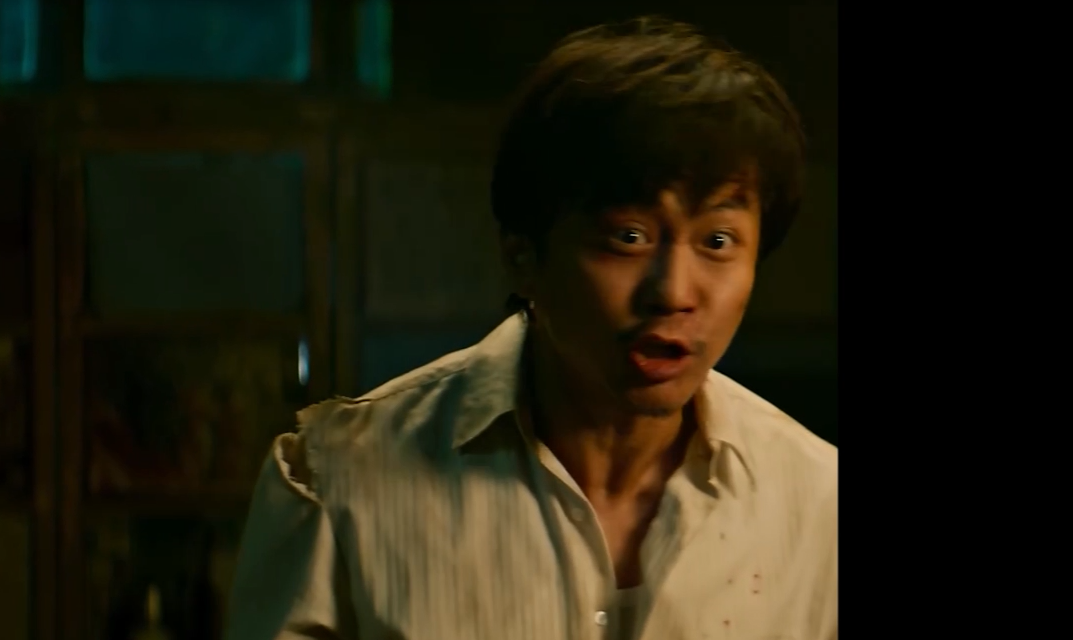} &
\includegraphics[width=0.32\linewidth]{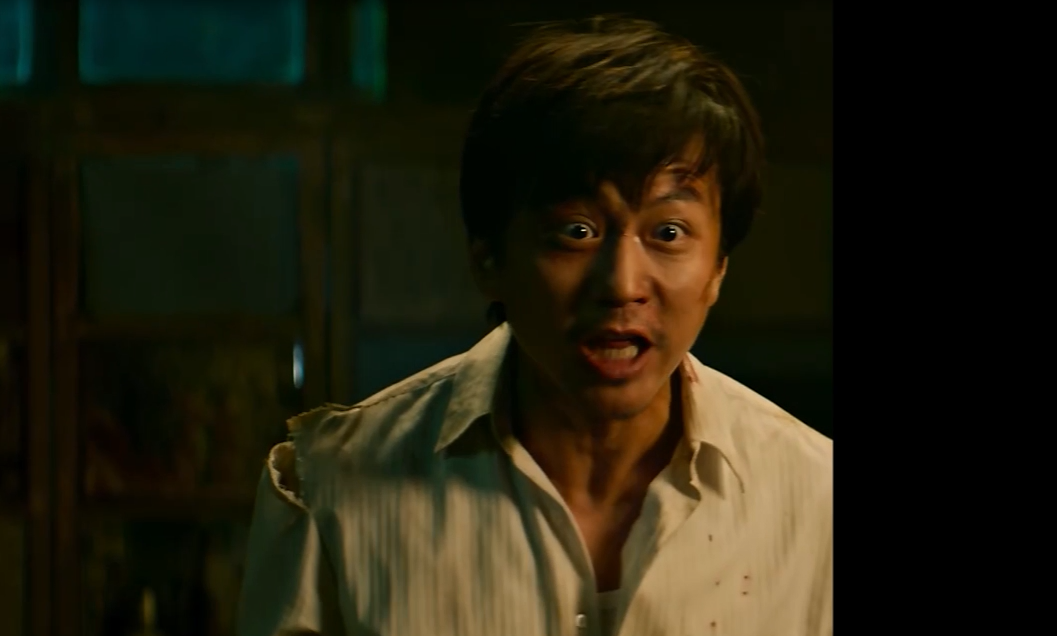} \\
{\scriptsize Frame-1} & {\scriptsize Frame-2} & {\scriptsize Frame-3}
\end{tabular}
\end{minipage}
\vspace{2pt}
\end{minipage}
} \\
\midrule

\textbf{Item} & \textbf{Baseline} & \textbf{CHASE} \\
\midrule

Prediction &
\texttt{label=P (1.0)} \xmark &
\texttt{label=N (-0.8)} \cmark \\
\midrule

Attention budget\newline\AttnLegend &
\AttnRow{L13}{0.9809}{0.0000}{0.0191}\newline
\AttnRow{L17}{0.9861}{0.0000}{0.0139}\newline
\AttnRow{L21}{0.9895}{0.0000}{0.0105}\newline
\AttnRow{L26}{0.9641}{0.0000}{0.0359}\newline
\AttnRow{L28}{0.9888}{0.0000}{0.0112}
&
\AttnRow{L13}{0.9347}{0.0000}{0.0653}\newline
\AttnRow{L17}{0.9389}{0.0000}{0.0611}\newline
\AttnRow{L21}{0.9415}{0.0000}{0.0585}\newline
\AttnRow{L26}{0.9216}{0.0000}{0.0784}\newline
\AttnRow{L28}{0.9515}{0.0000}{0.0485} \\
\midrule

Interpretation &
\multicolumn{2}{P{0.84\textwidth}}{%
\parbox[t]{0.84\textwidth}{%
\raggedright
\setlength{\emergencystretch}{2em}
\footnotesize
With missing audio, the decision relies on the TV pair.
The baseline predicts P(misaligned with GT=N), while CHASE shifts the attention budget toward more reliable visual evidence at late layers, recovering the GT-aligned negative prediction.
}} \\
\bottomrule
\end{tabular}

\vspace{-0.6em}
\caption{Missing case analysis with missing audio.}
\label{tab:missing_conflict_case_341}
\end{table*}

\end{document}